\documentclass[letterpaper]{article}
\usepackage{uai2019}
\usepackage[margin=1in]{geometry}

\usepackage[round]{natbib}
\usepackage{amsmath}
\usepackage{amsfonts}
\usepackage{centernot}

\usepackage[utf8]{inputenc}
\usepackage[english]{babel}
\usepackage[ruled,vlined,linesnumbered]{algorithm2e}

\usepackage{tikz}
\usetikzlibrary{arrows, decorations.markings}
\usetikzlibrary{decorations.pathreplacing}
\tikzstyle{vecArrow} = [thick, decoration={markings,mark=at position
	1 with {\arrow[semithick]{open triangle 60}}},
double distance=1.4pt, shorten >= 5.5pt,
preaction = {decorate},
postaction = {draw,line width=1.4pt, white,shorten >= 4.5pt}]
\tikzstyle{innerWhite} = [semithick, white,line width=1.4pt, shorten >= 4.5pt]

\usetikzlibrary{automata,positioning}

\newtheorem{theorem}{Theorem}
\newtheorem{property}{Property}

\usepackage{graphicx}
\newcommand{\bigCI}{\mathrel{\text{\scalebox{1.07}{$\perp\mkern-10mu\perp$}}}}
\newcommand{\nbigCI}{\centernot{\bigCI}}

\usepackage[utf8]{inputenc}
\usepackage[english]{babel}
\usepackage{nicefrac}
\newcommand{\kz}[1]{{\color{black} #1}}

\usepackage{times}

\title{Causal Discovery with General Non-Linear Relationships\\ Using Non-Linear ICA}

\author{
{ Ricardo Pio Monti}${^1}$, Kun Zhang$^2$,   \bf{Aapo Hyv\"{a}rinen}$^{1,3}$ \\
$^1$Gatsby Computational Neuroscience Unit, University College London, UK\\
$^2$Department of Philosophy, Carnegie Mellon University, USA\\
$^3$Department of Computer Science and HIIT, University of Helsinki, Finland
} 

\begin{document}

\maketitle

\begin{abstract}
We consider the problem of inferring 
causal relationships between two or more passively observed variables.
While the problem 
of such causal discovery 
has been extensively 
studied especially in the bivariate setting, the majority of current methods 
assume a linear causal relationship, and the few methods which 
consider non-linear relations usually 
make the assumption of additive noise. 
Here, we propose a framework through which we can perform causal 
discovery in the presence of general non-linear relationships.
The proposed method is based on recent progress in 
non-linear independent component analysis (ICA) and exploits the 
non-stationarity of observations 
in order to recover the underlying sources or latent disturbances. 
We show rigorously that in the case of bivariate causal discovery, such non-linear ICA can be used to infer causal direction via a series of independence tests. 
We further propose 
an alternative measure for inferring causal direction based on asymptotic 
approximations to the likelihood ratio, as well as an
extension to multivariate causal discovery. 
We demonstrate the capabilities of the proposed method via a series of 
simulation studies and conclude with an application to neuroimaging data. 
\end{abstract}

\section{INTRODUCTION}

Causal models play a fundamental role in modern scientific endeavor \citep{Pearl2009}. 
While randomized control studies 
are the gold standard, such an approach is unfeasible or unethical in many scenarios \citep{Spirtes2016}. 
Even when it is possible to run randomized control trials, the number  
of experiments 
required
may raise practical challenges  \citep{Eberhardt2005}.
Furthermore, big data sets publicly available on the internet often try to be generic and thus cannot be strongly based on specific interventions; a prominent 
example is the 
Human Connectome Project
which collects resting state fMRI 
data from over 500 subjects \citep{VanEssen2012}.
As such, it is both necessary and important to develop
\textit{causal discovery}
methods through which to uncover 
causal structure from (potentially large-scale)  
passively observed data.
Such data collected without  
explicit manipulation of certain variables is often termed
\textit{observational data}. 

The intrinsic appeal of causal discovery methods is that they allow us 
to uncover the underlying causal structure of complex systems, providing an 
explicit description 
of the underlying generative mechanisms. 
Within the context of machine learning 
causal discovery has also been shown to play an important role 
in many domains 
such as  
semi-supervised and transfer learning \citep{Scholkopf2012_causal, Zhang2013}, covariate shift 
and algorithmic fairness \citep{Kusner2017}. 
As a result, a wide range of methods have been 
proposed \citep{ Shimizu2006, Pearl2009, Hoyer2009, Zhangb, Peters2016, Zhang2017a}. 
However, many of the current methods rely on
restrictive assumptions regarding the nature of the causal relationships. 
For example, \cite{Shimizu2006}
assume linear causal models with non-Gaussian disturbances and demonstrate 
that independent component analysis (ICA) may be employed to uncover causal structure. 
\cite{Hoyer2009}  
provide an extension to 
non-linear causal models under the assumption of additive noise.

In this paper we propose a general method for bivariate causal discovery 
in the presence of general non-linearities. 
The proposed
method is able to uncover non-linear causal
relationships without requiring assumptions 
such as linear causal structure or additive noise.
Our approach
exploits a correspondence between
a non-linear ICA model and non-linear causal models,
and is specifically tailored for observational data which is collected 
across a series of distinct experimental conditions or regimes. 
Given such data, we seek to exploit 
the non-stationarity introduced via distinct experimental conditions in order to perform 
causal discovery. 
We demonstrate that 
if latent sources can be recovered via non-linear ICA, then  a series of 
independence tests may be employed to uncover causal structure. 
As an alternative to independence testing, we further propose a novel measure of non-linear  causal direction 
based on an asymptotic approximation to the likelihood ratio.

\section{PRELIMINARIES}
\label{sec_prelim}

In this section we 
introduce the class of causal models studied.
We also  
present an overview of non-linear ICA methods based on contrastive learning, upon which we base 
the proposed method.

\subsection{MODEL DEFINITION} 

Suppose we observe $d$-dimensional random variables $\textbf{X}= (X_1, \ldots, X_d)$ 
with joint distribution $\mathbb{P}(\textbf{X})$.  
The objective of causal discovery is to use the observed data, 
which give the empirical version of $\mathbb{P}(\textbf{X})$, to infer the associated 
causal graph
which describes the data generating procedure \citep{Pearl2009}.

A structural equation model (SEM) is here defined (generalizing the traditional definition) as a 
collection of $d$ 
structural equations:
\begin{equation}
\label{SEM_eq}
X_j = f_j ( \textbf{PA}_j, N_j), ~~ ~ j=1, \ldots ,d
\end{equation}
together with a joint distribution, $\mathbb{P}(\textbf{N})$, 
over disturbance (noise)
variables, $N_j$,  which are assumed to be mutually  independent. 
We write $\textbf{PA}_j$ to denote the parents of the variable $X_j$. 
The causal graph, $\mathcal{G}$, associated with 
a SEM in equation (\ref{SEM_eq})
is a graph
consisting of one node corresponding to each variable $X_j$;
throughout this work we assume $\mathcal{G}$ 
is a directed acyclic graph (DAG). 

While functions $f_j$ in equation (\ref{SEM_eq}) can be any
(possibly non-linear) functions, to date 
the causal discovery community has focused on 
specific special cases in order to obtain identifiability results as 
well as provide practical algorithms. 
Pertinent examples include:
 \textit{a)} 
the linear non-Gaussian acyclic model  (LiNGAM; \citeauthor{Shimizu2006}, 2006),
which assumes each $f_j$ is a linear function and the $N_j$ are non-Gaussian,
\textit{b)} the additive noise model (ANM; \citeauthor{Hoyer2009}, 2009), which assumes the noise is additive:
\begin{equation*}
\mathcal{S}_j : ~~ X_j = f_j ( \textbf{PA}_j) + N_j, ~~ ~ j=1, \ldots ,d,
\end{equation*}
and \textit{c)} the post-nonlinear causal model, which also captures possible non-linear distortion in the observed variables \citep{Zhangb}. 

The aforementioned approaches enforce strict  constraints on the functional class of the SEM. 
%
\kz{Otherwise, without suitable constraints on the functional class, then for any two variables one can always express one of them as a function of the other and independent noise \citep{Hyvarinen99}.
We are motivated to develop novel causal discovery methods 
which benefit from new identifiability results established from a different angle, in the context of 
general non-linear (and non-additive) relationships.}
A key component of our method exploits some recent advances in 
non-linear ICA, which we review next. 


\subsection{NON-LINEAR ICA VIA TCL}
\label{sec_nonLinICA}

We briefly outline the recently 
proposed Time Contrastive Learning (TCL) algorithm,
through which it is possible to demix (or disentangle) latent sources from observed non-linear mixtures; \kz{this algorithm provides hints as to the identifiability of causal direction between two variables in general non-linear cases under certain assumptions and is exploited in our causal discovery method.}
For further details we refer readers to \cite{Hyvarinen2016a}
but we also provide a brief review in Supplementary material \ref{sec_TCLreview}. 
We assume we observe $d$-dimensional data, $\textbf{X}$, which is generated according to a 
smooth and invertible non-linear mixture of 
independent latent variables $\textbf{S} = (S_1, \ldots, S_d)$. 
In particular, we have
\begin{equation}
\label{nonlin_ica_eq}
\textbf{X} = \textbf{f} ( \textbf{S} ).
\end{equation}
The goal of non-linear ICA  is then to recover $\textbf{S}$ from $\textbf{X}$. 

TCL, as 
introduced by \cite{Hyvarinen2016a},  is a method for non-linear ICA which is 
premised on the assumption that both latent sources
and observed data are non-stationary time series. Formally, they assume that while 
components $S_j$ are mutually independent, 
the distribution of each component is piece-wise stationary, implying 
they can be divided into 
non-overlapping time segments
such that their distribution varies across segments, indexed
by $e \in \mathcal{E}$.
In the basic case, 
the log-density of the $j$th latent source in segment $e$ is assumed to follow  
an exponential family distribution such that:
\begin{align}
\begin{split}
\label{gen_model} 
\mbox{log} ~p_e (S_j) = q_{j,0}(S_j) +  \lambda_{j} (e) q_{j}(S_j) - \mbox{log}~ Z(e),
\end{split}
\end{align}where $q_{j,0}$ is a stationary baseline and 
$q_j$
is a non-linear scalar function defining an exponential family for the 
$j$th source. (Exponential families with more than one sufficient statistic are also allowed.) 
The final term in equation (\ref{gen_model}) corresponds to a normalization constant.
It is important to note that parameters 
$\lambda_{j}(e)$ 
are functions of the
segment index, $e$, implying that the distribution of sources will vary across segments. 
It follows from equation (\ref{nonlin_ica_eq}) that 
observations $\textbf{X}$ may also be divided into non-overlapping segments indexed by $e \in \mathcal{E}$.
We write $\textbf{X}(i)$ to denote the $i$th observation and $C_i \in \mathcal{E}$ to denote its 
corresponding  
segment. 

TCL proceeds by defining a multinomial classification task, where we consider each original data 
 point
$\textbf{X}(i)$ as a data point to be classified, and the segment indices $C_i$ give the labels. 
Given the observations, $\textbf{X}$, together with 
the associated segment labels, $C$,  TCL can then be proven to
recover $\textbf{f}^{-1}$ 
as well as independent components, $\textbf{S}$, by learning to
classify the observations into their corresponding segments. 
In particular, TCL trains a deep neural network using multinomial logistic regression to
perform this classification task.
The network architecture employed consists of a feature 
extractor corresponding to the last hidden layer, denoted  $\textbf{h}( \textbf{X}(i); \theta)$ and 
parameterised by $\theta$, together with a final linear layer. 
The central Theorem on TCL is given in our notation as 
\begin{theorem}[\cite{Hyvarinen2016a}] \label{theorem_tcl}
	Assume the following conditions hold:
	\begin{enumerate}
		\item We observe data generated by independent sources according to equation (\ref{gen_model}) and mixed via invertible, smooth function $\normalfont \textbf{f}$ as stated in
		equation (\ref{nonlin_ica_eq}).
		\item We train a neural network consisting of a feature 
		extractor $\normalfont \textbf{h}(\textbf{X}(i); \theta)$  and a final linear layer (i.e., softmax classifier) to  
		classify each observation to its corresponding segment label, $C_i$. 
		We require the dimension of $\textbf{h}(\textbf{X}(i);\theta)$ be the same as  $\normalfont \textbf{X}(i)$. 
		\item The matrix $\normalfont \textbf{L}$ with elements 
		$\normalfont \textbf{L}_{e,i} = \lambda_{i}(e) - \lambda_{i}(1)$
		for $e=1, \ldots, E$ and   $i=1, \ldots, d$ has full  rank. 
	\end{enumerate}
	Then in the limit of infinite data, the outputs of the feature extractor 
	are equal to $\normalfont q( \textbf{S})$, up to an invertible linear transformation. 
\end{theorem}

Theorem \ref{theorem_tcl} states that 
we may perform non-linear ICA by training a neural network to 
classify  the 
segments
associated with each observation,
followed by 
linear ICA on the hidden representations, $\textbf{h}(\textbf{X}; \theta)$. 
This theorem provides identifiability of this particular non-linear ICA model, meaning that it is possible to recover the sources. 
This is not the case with many simpler attempts at non-linear ICA models \citep{Hyvarinen1999}, such as the case with a single segment  in the model above.

While Theorem \ref{theorem_tcl} provides identifiability for 
a particular non-linear ICA model, it requires a final linear unmixing of 
sources (i.e., via linear ICA). However, when sources 
follow the
piece-wise stationary distribution  detailed in equation (\ref{gen_model}), 
traditional linear ICA methods may not be appropriate
as sources will only be independent condition on the segment. For example, 
it is possible that exponential family parameters, $\lambda_j(e)$, 
are dependent across sources  (e.g., they may be correlated). 
This problem will be particularly pertinent when data is only collected over a 
reduced number of segments. 
As such, 
alternative linear ICA algorithms 
are required to effectively employ TCL in such a setting, as addressed in Section~\ref{SM_linICAmodel}.

\section{NON-LINEAR CAUSAL DISCOVERY VIA NON-LINEAR ICA}
\label{sec_PropMethod}

In this section we outline the 
proposed method for causal discovery over bivariate data, which we term
\textbf{Non}-linear \textbf{S}EM \textbf{E}stimation using \textbf{N}on-\textbf{S}tationarity (NonSENS).
We begin by providing an intuition for the proposed method in Section \ref{sec_SEM_to_ICA}, which is
based on the connection between non-linear ICA and 
non-linear SEMs.
In Section \ref{SM_linICAmodel} we propose a novel linear ICA algorithm which \kz{complements TCL} for the purpose of causal discovery, particularly
in the presence of observational data with few segments.
Our method is formally detailed in 
Section \ref{sec_NonSENS_indepTest}, 
which also contains  
a proof of 
identifiability. Finally in Section \ref{sec_assumeCause} we present an alternative measure of 
casual direction based on asymptotic approximations to the likelihood ratio of 
non-linear causal models.



\subsection{RELATING SEM TO ICA}
\label{sec_SEM_to_ICA}
We assume we observe bivariate data 
$\textbf{X}(i) \in \mathbb{R}^2$
and write 
${X}_{1}(i) $ and  ${X}_{2}(i)$ to denote the
first and second entries of $\textbf{X}(i)$ respectively. 
We will omit the $i$ index whenever it is clear from context. 
%
Following the notation of \cite{Peters2016},
we further assume 
data is available over a set of 
distinct environmental conditions $\mathcal{E} = \{1, \ldots, E\}$. 
As such, each $\textbf{X}(i)$ is allocated to an experimental condition and 
denote by $C_i \in \mathcal{E}$ 
 the experimental condition under which the $i$th observation was generated.
Let $n_e$ be the number of observations within each experimental condition
such that $n_{tot}= \sum_{e \in \mathcal{E} }n_e$. 

\label{sec_TCL_testing}

\begin{figure}[t!]
	\begin{center}
		\includegraphics[width=.45\textwidth]{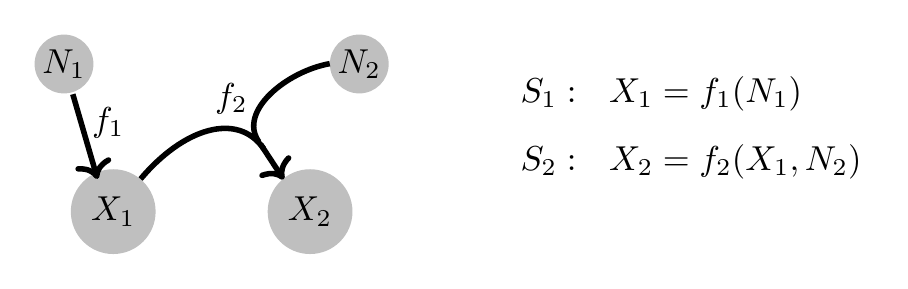}
	\end{center} \vspace{-4mm}
	\caption{Visualization of DAG, $\mathcal{G}$, associated with the SEM
		in equations (\ref{bivariate_eq1}) and (\ref{bivariate_eq2}). }
	\label{DAG_Fig}
\end{figure}

The objective of the 
proposed method is to uncover the causal direction between
${X}_1$ and ${X}_2$. 
Without loss of generality, we explain the basic logic. Suppose that ${X}_1 \rightarrow {X}_2$, such that 
the associated SEM is of the form:
\begin{align}
\label{bivariate_eq1}
{X}_1(i) &= f_1( N_1(i) ),\\
{X}_2(i) &= f_2({X}_1(i), N_2(i)),
\label{bivariate_eq2}
\end{align}
where $N_1, N_2$ are 
latent disturbances whose distribution is also
assumed to vary across experimental conditions.
The DAG associated with equations (\ref{bivariate_eq1}) and (\ref{bivariate_eq2}) is shown in Figure \ref{DAG_Fig}.
Fundamentally, the proposed NonSENS algorithm exploits the 
correspondence between the non-linear ICA model described in Section \ref{sec_nonLinICA}
and non-linear SEMs. 
This correspondence is formally stated as follows:
observations generated according to the piece-wise 
stationary non-linear ICA model of
equations (\ref{nonlin_ica_eq})
and (\ref{gen_model})
will follow a (possibly non-linear) SEM where each 
disturbance variance, $N_j$, corresponds to a latent source, $S_{\pi(j)}$, 
and each structural equation, $f_j$, will correspond to an entry in 
the smooth, invertible funciton $\textbf{f}$. We note that due to the permutation indeterminacy 
present in ICA 
each disturbance variable, $N_j$, will only be identifiable up to some permutation 
$\pi$ of the set $\{1,2\}$. 

The proposed method consists of a two-step procedure.
First, it seeks to recover latent disturbances 
via non-linear ICA. 
Given estimated latent disturbances, the following property 
highlights how 
we may employ 
statistical independencies between 
observations and estimated sources in order to infer the causal structure:
\begin{property}
	\label{Prop_DAG_indep} Assume the true causal structure follows equations  (\ref{bivariate_eq1}) and (\ref{bivariate_eq2}), as depicted in 
	Figure \ref{DAG_Fig}. Then, assuming each 
	observed variable
	is statistically dependent on its latent disturbance (thus avoiding degenerate cases),
	it follows that $X_1 \bigCI  N_2 $ while 
	$ X_1 \nbigCI N_1$ and $ X_2 \nbigCI N_1$ as well as $ X_2 \nbigCI N_2$. \footnote{We note that the property that effect is dependent on its direct causes typically holds, although one may construct specific examples (with discrete variables or continuous variables with complex causal relations) in which effect is independent from its direct causes. In particular, if faithfulness is assumed~\citep{Spirtes2000}, the above property clearly holds.}
\end{property}

Property \ref{Prop_DAG_indep} highlights the relationship between observations $\textbf{X}$
and  latent sources, $\textbf{N}$, and provides some insight into how
a non-linear ICA method, together with 
independence testing, could be 
employed to perform bivariate causal discovery.  This is formalized in
Section \ref{sec_NonSENS_indepTest}.


\subsection{A LINEAR ICA ALGORITHM FOR PIECE-WISE STATIONARY SOURCES}
\label{SM_linICAmodel}

Before proceeding, we have to improve the non-linear ICA theory of \cite{Hyvarinen2016a}.
Assumptions 1--3 of  Theorem \ref{theorem_tcl} for TCL guarantee that 
the feature extractor, $\textbf{h}(\textbf{X}; \theta)$, will 
recover a linear mixture of latent independent sources. 
As a result, 
applying a linear unmixing method to the final representations, $\textbf{h}(\textbf{X}; \theta)$,
will allow us to recover latent disturbances. 
However, the use of ordinary linear ICA to unmix $\textbf{h}(\textbf{X}; \theta)$
 is premised on the assumption that latent sources 
are independent. This is not necessarily guaranteed when sources follow the
ICA model presented in 
equation (\ref{gen_model}) with a fixed number of segments. For example, it is possible that
parameters, $\lambda_j(e)$, are correlated across segments. 
We note that this is not a problem when the number of segments increases
asymptotically and 
parameters, $\lambda_j(e)$, are 
assumed to be randomly generated, as stated in Corollary 1 of \cite{Hyvarinen2016a}. 

In order to address this issue, we propose an alternative linear
ICA algorithm 
to be employed in the final stage of TCL, through which to accurately recover
latent sources in the presence of a small number of segments.


The proposed linear ICA algorithm explicitly models latent sources as 
following the piece-wise stationary distribution specified in equation (\ref{gen_model}). 
We write $\textbf{Z}(i)\in \mathbb{R}^d$ to denote the $i$th observation, 
generated as a linear mixtures of sources:
$\textbf{Z}(i) = \textbf{A}\textbf{S}(i)$, where $\textbf{A} \in \mathbb{R}^{d \times d}$ is a square mixing matrix. 
Estimation of parameters proceeds via score matching \citep{Hyvarinen2006}, which yields an objective
function of the following form:
{\scriptsize
\begin{align*}
\label{SM_obj}
& J = \sum_{e \in \mathcal{E}} \sum_{j=1}^d \lambda_j (e) \frac{1}{n_e} ~\sum_{C_i=e} q_j'' (\textbf{w}^T_j \textbf{Z}(i) ) \\&+ 
\frac{1}{2} \sum_{e \in \mathcal{E}} \sum_{j,k=1}^d \lambda_k(e) \lambda_j(e) \textbf{w}_k^T \textbf{w}_j
\frac{1}{n_e} \sum_{C_i=e} q_k'(\textbf{w}^T_k \textbf{Z}(i)) q_j'(\textbf{w}^T_j \textbf{Z}(i) ),
\end{align*}}where $\textbf{W}\in \mathbb{R}^{d \times d}$ denotes the unmixing matrix and 
$q_j'$ and $q_j''$ denote the first and second derivatives of 
the non-linear scalar functions introduced in equation (\ref{gen_model}).
Details and results are provided in Supplementary \ref{supp_linearICA_SM}, where the 
proposed method is  shown to outperform  
 both FastICA and Infomax ICA, as well as 
 the joint diagonalization method of \cite{Pham2001a}, which is explicitly tailored for 
 non-stationary sources.

\subsection{CAUSAL DISCOVERY USING INDEPENDENCE TESTS}
\label{sec_NonSENS_indepTest}

Now we give the outline 
NonSENS. NonSENS performs causal 
discovery by combining
Property \ref{Prop_DAG_indep} with 
a non-linear ICA algorithm. 
Notably, we employ TCL, described in Section \ref{sec_nonLinICA},  with the 
important addition that the final linear unmixing of the hidden representations,
$\textbf{h}(\textbf{X}; \theta)$, is performed using the 
objective given in Section \ref{SM_linICAmodel}. 
The proposed method is summarized as follows: 
\begin{enumerate}
	\item \begin{enumerate}
		\item Using TCL, train a deep neural network with 
		feature extractor $\textbf{h}(\textbf{X}(i); \theta)$ 
		to accurately classify each observation $\textbf{X}(i)$ according to its segment label $C_i$.
		\item Perform linear unmixing of $\textbf{h}(\textbf{X}; \theta)$  using the algorithm 
		presented in Section \ref{SM_linICAmodel}.
	\end{enumerate}
\item Perform the four tests listed in Property 1, and conclude a causal 
effect in the case where
there is evidence to reject the null hypothesis in three of the tests and 
only one of the tests fails to reject the null. 
The variable
for which the null hypothesis was not rejected is
the cause variable. 
\end{enumerate}


%
%
%
%
Each test is run at a pre-specified significance level, $\alpha$, and Bonferroni 
corrected in order to control the 
family-wise error rate.
Throughout this work 
we employ HSIC as a test for statistical independence \citep{Gretton2005b}.
%
Theorem \ref{TCL_causalRecov_prop} 
formally states the assumptions
and identifiability properties of the
proposed method.

\begin{theorem}
	\label{TCL_causalRecov_prop}
	Assume the following conditions hold:
	\begin{enumerate}
		\item We observe bivariate data $\textbf{X}$ which has been generated 
		from a non-linear SEM with smooth non-linearities and no hidden confounders. 
		\item Data is available 
		over at least three distinct experimental
		conditions 
		and latent disturbances, $N_j$, are generated according 
		to equation (\ref{gen_model}).
		\item We employ TCL, with a sufficiently deep neural network as the feature extractor,
		followed by linear ICA (as described in Section \ref{SM_linICAmodel}) on hidden representations 
		to recover the latent sources. 
		\item We employ an independence test 
		which can capture any type of departure from independence, for example HSIC, with Bonferroni correction and significance level $\alpha$. 
	\end{enumerate}   
	Then 
	in the limit of infinite data  
	the proposed method will  identify the cause variable
	with probability $1 - \alpha$. 
\end{theorem} 

\kz{See Supplementary \ref{supp_Proof_Theorem2} for a proof. Theorem \ref{TCL_causalRecov_prop} extends previous identifiability 
results relying on constraints on functional classes (e.g., ANM in \cite{Hoyer2009}) to the 
domain of arbitrary non-linear models, under further assumptions on nonstationarity of the given data.}


\subsection{LIKELIHOOD RATIO-BASED MEASURES OF CAUSAL DIRECTION}
\label{sec_assumeCause}
While independence tests are widely used in causal discovery, they may not be statistically optimal for deciding causal direction.
In this section, we further propose a novel measure of causal direction which is based 
on the likelihood ratio under non-linear causal models, and which thus is likely to be more efficient. 
%

The proposed measure can be seen as the extension of linear measures of causal direction, such as
those proposed by \cite{Hyvarinen2013}, to the domain of non-linear SEMs. 
Briefly, 
\cite{Hyvarinen2013}
consider the likelihood ratio between two candidate models of causal influence: $X_1 \rightarrow X_2$
or $X_2 \rightarrow X_1$. The log-likelihood ratio is then defined as the difference in log-likelihoods 
under each model:
\begin{equation}
\label{LR_definition}
R = L_{1\rightarrow 2} - L_{2 \rightarrow 1}
\end{equation}
where we write $ L_{1\rightarrow 2}$ to denote the 
log-likelihood under the  assumption that $X_1$ is the causal variable
and $ L_{2\rightarrow 1}$ for the alternative model.
Under the assumption that $X_1 \rightarrow X_2$, 
it follows that the underlying SEM is of the form described in 
equations (\ref{bivariate_eq1}) and (\ref{bivariate_eq2}).
The log-likelihood for a single data point may thus be written as
\begin{align*}
\scriptsize 
 L_{1\rightarrow 2} 
&= \log P_{X_1}(X_1) + \log P_{X_2 | X_1}( X_2 | X_1).
\end{align*}
Furthermore, in the context of linear causal models 
we have that equations (\ref{bivariate_eq1}) and (\ref{bivariate_eq2}) define a bijection between $N_2$ and $X_2$
whose Jacobian has unit determinant, such that 
the log-likelihood can be expressed as:
\begin{align*}
 L_{1\rightarrow 2}
 = \log P_{X_1}(X_1) + \log P_{N_2}(N_2). 
\end{align*}
In the asymptotic limit we can take the expectation of log-likelihood, and the log-likelihood converges to:
\begin{equation}
\label{neg_entropy_lin}
\mathbb{E} [L_{1 \rightarrow 2}] = -H(X_1)  - H( N_2 )
\end{equation}
where $H(\cdot)$ denotes the differential entropy. 
\cite{Hyvarinen2013} note that the benefit of equation (\ref{neg_entropy_lin})
is that only univariate approximations 
of the differential entropy are required.
%
%
In this section we seek to derive equivalent measures for causal direction without
the assumption of linear causal effects. 
Recall that after training via TCL, we obtain an estimate 
of $\textbf{g} = \textbf{f}^{-1}$ 
which is parameterized by a deep neural network. 

In order to compute the log-likelihood, $L_{1\rightarrow 2}$, 
we consider the following change of variables:
	\begin{equation*}
\label{change_of_var}
\binom{X_1}{N_{2}} =  \mathbf{\tilde g} \binom{ X_1 }{X_2} = \binom{ X_1 }{ \mathbf{g}_{2}(X_1,X_2) }
\end{equation*}
where 
we note that 
$\mathbf{g}_{2}: \mathbb{R}^2 \rightarrow \mathbb{R} $  refers to the  second component 
of $\mathbf{g}$.
Further, we note that the the mapping $\mathbf{\tilde g}$ only applies the identity 
to the first element, thereby leaving  
$X_1$ unchanged. 
%
%
%
%
%
Given such a change of variables, we may evaluate the log-likelihood 
as follows:
	\begin{align*}
	\small 
L_{1\rightarrow 2}
&= \log p_{X_1} (X_1) + \log p_{N_{2}}(N_{2}) + \log | \mbox{det}~ \mathbf{J \tilde g}|,
\end{align*}
where $\mathbf{J \tilde g}$ denotes the Jacobian of $\mathbf{\tilde g}$, as we have $X_1 \bigCI N_{2}$ by construction under the assumption that $X_1 \rightarrow X_2$. 

Due to the particular choice of $\mathbf{ \tilde g}$, we are able to easily evaluate 
the Jacobian, which can be expressed as:
%
%
%
%
	\begin{equation*}
\label{Jacobian_1}
\mathbf{J \tilde g} = \left ( 
\begin{matrix} 
\frac{\partial \mathbf{\tilde g}_1}{\partial X_1} & \frac{\partial \mathbf{\tilde g}_1}{\partial X_2} \\ 
\frac{\partial \mathbf{\tilde g}_2}{\partial X_1} & \frac{\partial \mathbf{\tilde g}_2}{\partial X_2}  \\ 
\end{matrix}
\right ) =  \left ( \begin{matrix}
1 & 0 \\
\frac{\partial { \mathbf{g}_{2}}}{\partial X_1} & \frac{ \partial {\mathbf{g}_{2}}}{\partial X_2}  \\ 
\end{matrix}\right ).
\end{equation*}
As a result, the determinant can be directly evaluated as $ \frac{ \partial {\mathbf{g}_{2}}}{\partial X_2}$. Furthermore, since $\textbf{g}_{ 2}$ is parameterized by a deep network, we can directly evaluate its derivative with respect to $X_2$. 
This allows us to directly evaluate the log-likelihood 
of $X_1$ being the causal variable as:
\begin{equation*}
L_{1 \rightarrow 2}  
= \log p_{X_1} (X_1) + \log p_{N_{2}}(N_{2}) + \log \left |  \frac{\partial {\mathbf{g}_{2}}}{\partial X_2} \right |.
\end{equation*} 
Finally, we consider the asymptotic limit and obtain the non-linear 
generalization of equation (\ref{neg_entropy_lin}) as:
{ 
\begin{align*}
\label{neg_entropy_nonlin}
\mathbb{E} [L_{1 \rightarrow 2} ]
 = &-H(X_1) - H(N_{2})
   +  \mathbb{E}\left [\log \left | \frac{\partial {\mathbf{g}_{2}}}{\partial X_2}\right | \right ]. 
\end{align*}}In practice we use the sample mean instead of the expectation.



One remaining issue to address is the permutation invariance of estimated sources \kz{(note this this permutation is not about the causal order of the observed variables)}. 
We must consider both permutations $\pi$ of the set $\{1,2\}$. 
In order to resolve this issue, we note that
if the true permutation is $\pi = (1,2)$ 
then assuming $X_1 \rightarrow X_2$ we have
$\frac{\partial \mathbf{g}_1}{\partial X_2}=0$ while 
$\frac{\partial \mathbf{g}_2}{\partial X_2}\neq 0$. This is because $\mathbf{g}_1$ unmixes 
observations to return the latent disturbance for causal variable, $X_1$, and is therefore not
a function of $X_2$. The converse is true if the permutation is $\pi=(2,1)$. Similar reasoning can be employed 
for the reverse model: $X_2 \rightarrow X_1$. 
As such, we  propose to select the permutation as follows:
{ \small
\begin{equation*}
    \pi^* = \underset{\pi}{\operatorname{argmax}} \left \{   \mathbb{E}\left [ \log \left |\frac{\partial {\mathbf{g}_{\pi(2)}}}{\partial X_2}\right |\right ] + \mathbb{E} \left [\log \left |\frac{\partial {\mathbf{g}_{\pi(1)}}}{\partial X_1} \right | \right ] \right \}.
\end{equation*}}For a given permutation, $\pi$,
we may therefore compute the likelihood ratio 
in equation (\ref{LR_definition}) as:
{ \small
\begin{align*}
    R = -H(X_1) - H(N_{\pi(2)})
   &+ \mathbb{E} \left [\log \left |\frac{\partial {\mathbf{g}_{\pi(2)}}}{\partial X_2} \right | \right ]\\ 
    +H(X_2) + H(N_{\pi(1)}) &- \mathbb{E} \left [ \log \left |\frac{\partial {\mathbf{g}_{\pi(1)}}}{\partial X_1}\right |  \right].
\end{align*}
}If $R$ is positive, we conclude that $X_1$ is the causal 
variable, whereas if $R$ is negative $X_2$ is reported as the causal variable. 
When computing the differential entropy, we employ the approximations 
described in \cite{Kraskov2004}. We note that such approximations require variables to be 
	standardized; in the case of latent variables this can be achieved by defining a further  
	change of variables corresponding to a standardization.  

Finally, we note that the likelihood ratio presented above can be connected to 
the independence measures employed in Section \ref{sec_NonSENS_indepTest} 
when mutual information is used a measure of statistical dependence.
In particular, we have
\begin{equation}
    R =  -I( X_1, N_{\pi(2)}) + I (X_2, N_{\pi(1)}),
    \label{LR_ind_eq}
\end{equation}
where $I(\cdot, \cdot)$ denotes the mutual information between two variables. We provide a full derivation in 
Supplementary \ref{LR_to_indTest_supp}. This result serves to connect the proposed likelihood 
ratio 
to independence testing methods for causal discovery which use mutual information.


\subsection{EXTENSION TO MULTIVARIATE DATA} 
\label{sec_multivariateCD}

It is not straightforward to extend NonSENS to multivariate cases. 
Due to the permutation invariance of sources, independence testing would require
$d^2$ tests, where $d$ is the number of variables, 
leading to a significant drop in power after  Bonferroni correction. Likewise, the likelihood ratio test inherently considers only two variables. 


Instead, we propose to extend to proposed method to the domain of multivariate 
causal discovery 
by employing it in conjunction with a traditional constraint based
method such as the PC algorithm, as in \cite{Zhangb}.
Formally, the PC algorithm is first employed to estimate 
the skeleton and orient as many edges as possible. Any remaining undirected edges
are then
directed using either proposed bivariate method. 

\subsection{RELATIONSHIP TO PREVIOUS METHODS} 

NonSENS is closely related to 
linear ICA-based methods as described in \cite{Shimizu2006}. 
However,
there are  important differences: 
LiNGAM focuses exclusively on linear causal models whilst NonSENS is specifically designed to recover arbitrary non-linear 
causal structure. 
Moreover,
the proposed method is mainly designed for bivariate causal discovery whereas the original LiNGAM method can easily
perform multivariate causal discovery by permuting the 
estimated ICA unmixing matrix. 
In this sense NonSENS is more closely aligned to the 
Pairwise LiNGAM method \citep{Hyvarinen2013}.

\cite{Peters2013}  propose a non-linear causal discovery method 
named regression and subsequent independence test (RESIT)
which is able to recover the causal structure under the assumption of an 
additive noise model. 
RESIT essentially shares the same underlying idea as NonSENS, with the 
difference being that 
it estimates latent disturbances 
via non-linear regression, as opposed to via 
non-linear ICA. 
Related to the RESIT algorithm is the 
Regression Error Causal Inference (RECI) algorithm  \citep{Blobaum2018}, which proposes measures of causal direction based on the magnitude of
(non-linear)
regression errors. Importantly, both of those methods restrict the non-linear relations to have additive noise.


Recently several methods have been proposed which seek to exploit 
non-stationarity in order to perform causal discovery.
For example, \cite{Peters2016}  propose to leverage the 
invariance of 
causal models under covariate shift in order to recover the true causal structure. 
Their method, termed Invariant Causal Prediction (ICP), is
tailored to the setting where data is collected across a variety of  experimental regimes, similar to ours. However, their main results, including identifiability are in the linear or additive noise settings.

In contrast, \cite{Zhang2017a} proposed a  method, termed CD-NOD, for 
causal discovery from heterogeneous, multiple-domain data or non-stationary data, which allows for general non-linearities. Their method thus solves a problem very similar to ours, although with a very different approach. 
Their  method
accounts for non-stationarity, which manifests itself via changes in the causal modules,
via the introduction of an surrogate variable 
representing the  domain or time index into the causal DAG.
Conditional independence testing is  employed to 
recover the skeleton over the augmented DAG, and their method does not produce an estimate of the SEM to represent the causal mechanism. 

\section{EXPERIMENTAL RESULTS}
\label{sec_ExpRes}

In order to demonstrate the capabilities of the proposed method we consider a 
series of experiments on synthetic data as well as real neuroimaging data. 

\subsection{SIMULATIONS ON ARTIFICIAL DATA}
In the implementation of the proposed method we 
employed 
deep neural networks of varying depths as feature extractors. 
All  networks 
were trained 
on cross-entropy loss 
using stochastic gradient descent.
In the final linear unmixing required by TCL, we employ the 
linear ICA model described in Section \ref{SM_linICAmodel}. 
For independence testing, we employ HSIC with a Gaussian kernel.
All tests are run at the $\alpha = 5\%$ level and Bonferroni corrected.

We benchmark the performance of the 
NonSENS algorithm
against several 
state-of-the-art methods. 
As a measure of performance against linear methods we 
compare against LiNGAM. In particular, 
we compare performance to 
DirectLiNGAM \citep{Shimizu2011}.
In order to highlight the need for non-linear ICA methods, we also consider the performance of the 
proposed method where linear ICA is employed to estimate latent disturbances; 
We
refer to this baseline as Linear-ICA NonSENS. 
We further compare against the RESIT method of 
\cite{Peters2013}.
Here we employ Gaussian process regression to estimate non-linear effects and 
HSIC as a measure of statistical dependence. 
Finally, we also compare against the CD-NOD method of \cite{Zhang2017a}
as well as the RECI 
method presented in 
\cite{Blobaum2018}. For the latter, we employ Gaussian process regression and note that 
this method assumes the presence of a  causal effect, and is therefore only  included
in some experiments. 
We provide a description of each of the methods in the 
Supplementary material 
\ref{app_baselines}. 


We generate synthetic data 
from the non-linear ICA model detailed in 
Section \ref{sec_nonLinICA}. 
Non-stationary disturbances, $\textbf{N}$, 
were randomly generated by simulating 
Laplace random variables with distinct variances in each segment.
For the non-linear mixing function we employ a 
deep neural network (``mixing-DNN") with randomly generated weights such that:
\begin{align}
\textbf{X}^{(1)} &= \textbf{A}^{(1)} \textbf{N}, \label{lingam_eq}\\
\textbf{X}^{(l)} &= \textbf{A}^{(l)} {f} \left ( \textbf{X}^{(l-1)}, \right ) \label{general_eq}
\end{align}
where we write $\textbf{X}^{(l)}$ to denote the activations at the $l$th layer
and $f$ corresponds to the leaky-ReLU activation function which is applied element-wise.
We restrict 
matrices $\textbf{A}^{(l)}$ to be lower-triangular in order to  introduce
acyclic causal relations.
In the special case of multivariate causal discovery, 
we follow  \cite{Peters2013} and randomly
include edges with a probability of $\nicefrac{2}{d-1}$, implying that the expected number of 
edges is $d$. We present experiments for $d=6$ dimensional data.  
Note that equation (\ref{lingam_eq}) is a LiNGAM. For depths $l \geq 2$,
equation (\ref{general_eq}) generates  data with non-linear causal structure. 


Throughout  experiments we 
vary the following factors: the number of distinct 
experimental conditions (i.e., distinct segments, $|\mathcal{E}|$),
the number of observations per segment, $n_e$, 
as well as the depth, $l$, of the mixing-DNN. 
%
%
%
In the context of bivariate causal discovery we measure  
how frequently each method is able to correctly identify the cause variable.
For multivariate causal discovery we consider the 
$F_1$ score, which serves to quantify the agreement between estimated and true 
DAGs. 


\begin{figure*}[ht!]
	\centering
	\includegraphics[width=.87\textwidth]{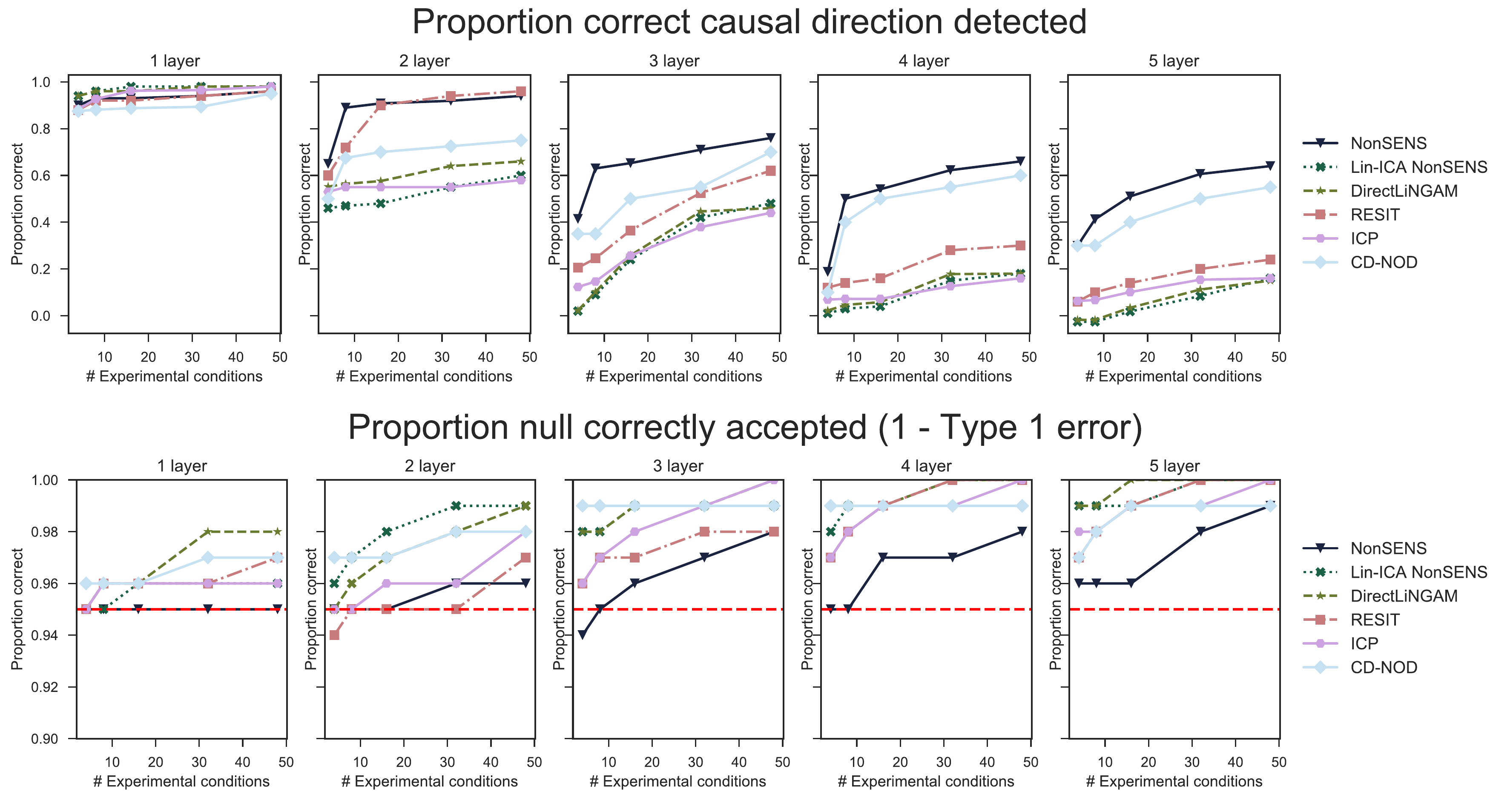}
	\caption{Experimental results 
		indicating performance as we increase the number of experimental conditions, $|\mathcal{E}|$, 
		whilst keeping the number of observation per condition fixed at $n_e = 512$.
		Each horizontal panel plots results for varying depths of the mixing-DNN, ranging from $l=1, \ldots 5$. 
		The top panels show the proportion of times the correct cause variable is identified when a causal effect exists. The bottom panels considers data where 
		no acyclic causal structure exists
		($\textbf{A}^{(l)}$ are not lower-triangular) and reports the 
		proportion of times no causal effect is correctly reported. 
		The dashed, horizontal 
		red line indicates the theoretical $(1-\alpha)\%$ true negative rate. 
		For clarity we omit the standard errors, but we note that they were  small in magnitude (approximately $2-5\%$). 
	}
	\label{fig:SimResultsIncSeg}
\end{figure*}

\begin{figure*}[ht!]
	\centering
	\includegraphics[width=.87\textwidth]{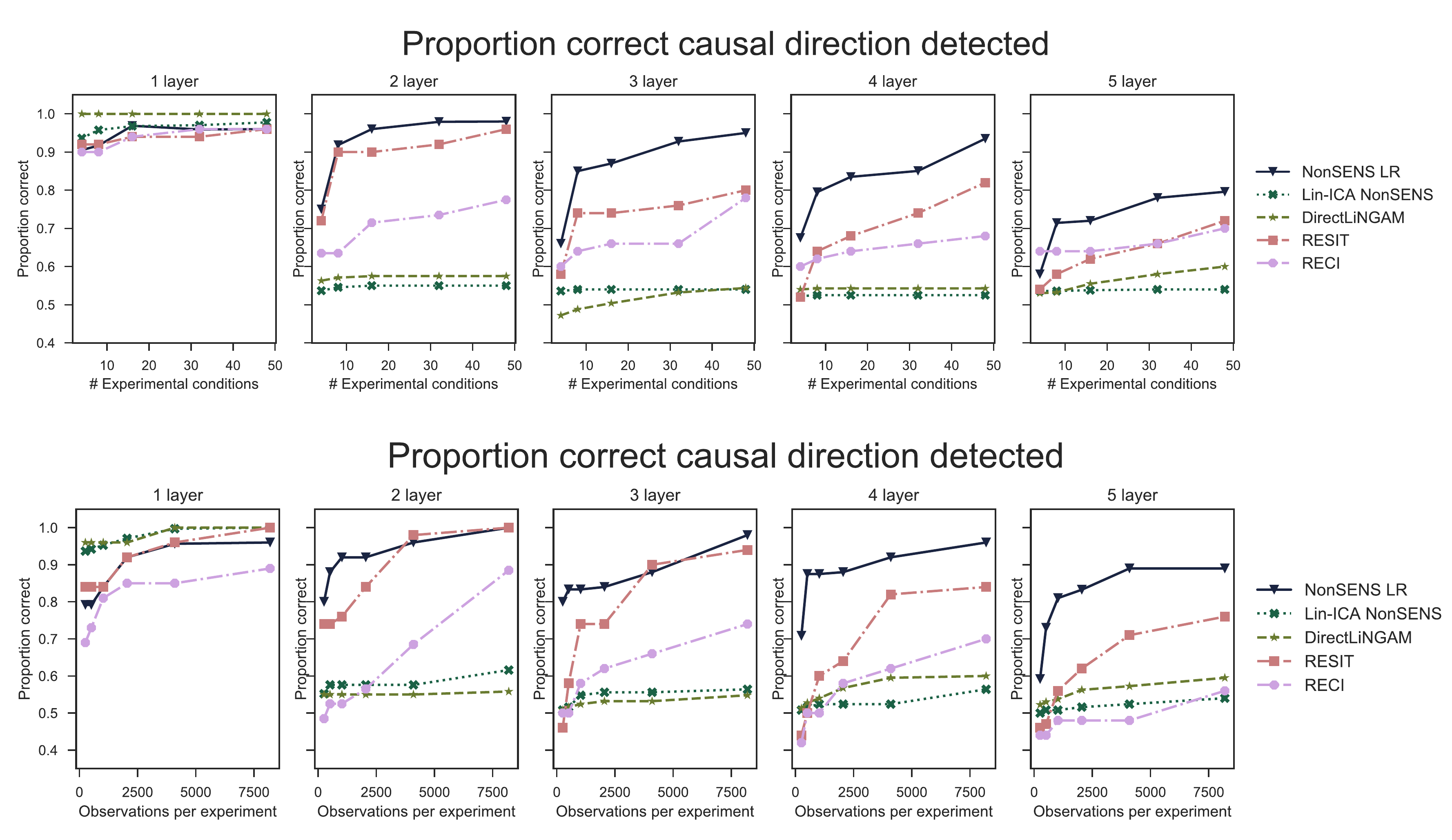}
	\caption{Experimental results 
		visualizing  performance under the assumption that a causal effect exists. 
		This reduces the bivariate causal discovery problem to recovering the causal 
		ordering over $X_1$ and $X_2$. 
		The top panel considers an increasing number of experimental conditions 
		whilst the bottom panel shows results when we vary the number of 
		observations within a fixed number of experimental conditions, $|\mathcal{E}|=10$.
		Each horizontal plane plots results for varying depths of the mixing-DNN, ranging from $l=1, \ldots, 5$.
	}
	\label{fig:SimResultsAssumeCause}
\end{figure*}

Figure \ref{fig:SimResultsIncSeg} shows results for bivariate causal discovery
as the number of distinct experimental conditions, $|\mathcal{E}|$, increases
and the  number of observations within each condition was fixed at $n_e = 512$. 
Each horizontal panel shows the results as the depth of the 
mixing-DNN increased from $l=1$ to $l=5$.
The top panels show the proportion of times the correct cause variable was identified across
100 independent simulations.  
In particular, the first top panel corresponds 
to
linear causal dependencies.
As such, 
all methods are able to accurately recover the true cause variable.
However, as the depth of the mixing-DNN increases, the causal dependencies become 
increasingly non-linear and the performance of all methods deteriorates. 
While we attribute this drop in performance  to the increasingly non-linear 
nature of causal structure, we  note that
the NonSENS algorithm 
is able to 
out-perform all alternative methods. 

The bottom panels of Figure \ref{fig:SimResultsIncSeg} shows the results when
no directed acyclic causal structure is present. Here data was generated such that $\textbf{A}^{(l)}$ was not lower-triangular.
In particular, we set the off-diagonal entries of  $\textbf{A}^{(l)}$ to be equal and non-zero,
resulting in cyclic causal structure. 
In the context of such data, we would expect all methods to report that the 
causal structure is 
inconclusive $95\%$ of the time, as all tests are Bonferroni corrected
at the $\alpha=5\%$ level. 
The bottom panel of Figure \ref{fig:SimResultsIncSeg} show the proportion of times 
the causal structure is correctly reported as inconclusive.
The results indicate that all methods are overly conservative in their testing, and become 
increasingly conservative as the depth, $l$, increases. 

We also consider the performance of all algorithms in the context of 
a fixed number of 
experimental conditions, $|\mathcal{E}|=10$, 
and an increasing number of observations per condition, $n_e$. 
These results, presented in 
Supplementary \ref{extra_experiments_supp}, demonstrate that the 
proposed method continues to perform competitively in such a scenario. 

\begin{figure}[t!]
	\centering
	\includegraphics[width=.475\textwidth]{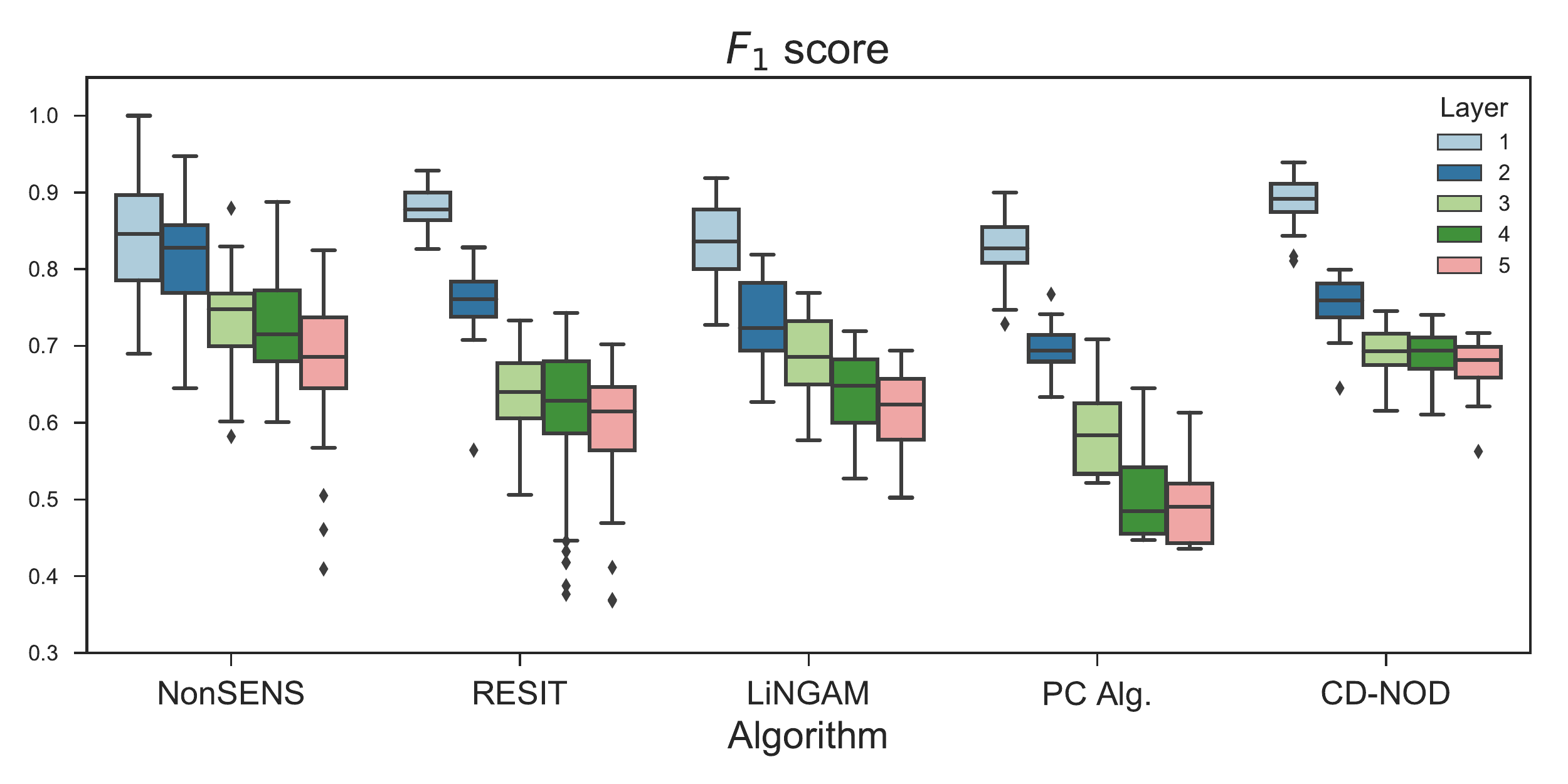} \vspace{-3mm}
	\caption{$F_1$ score for multivariate causal discovery over  
		$6$-dimensional data. 
		For each algorithm, we plot the $F_1$ scores as we vary the 
		depth of the mixing-DNN from $l=1, \ldots, 5$.
		Higher $F_1$ scores indicate better performance. 
	}
	\label{fig:SimResultsGeneralCase}
\end{figure}

Furthermore, we also consider the scenario where a causal effect is assumed to exist.
In such a scenario, we employ 
the likelihood ratio approach described in Section \ref{sec_assumeCause}.
 In the case of algorithms such as 
 RESIT we compare $p$-values in order to determine direction. 
The results for these experiments are shown in Figure \ref{fig:SimResultsAssumeCause}.
The top panels show results as the number of 
experimental conditions, $|\mathcal{E}|$, increases. As before, we fix the number of 
observations per condition to $n_e=512$. 
The bottom panels show results for a fixed number of experimental conditions
$|\mathcal{E}|=10$, as we increase the number of observations per condition.
We note that the proposed measure of causal direction is shown to outperform alternative algorithms. 
Performance in Figure \ref{fig:SimResultsAssumeCause} appears 
significantly higher than that shown in Figure \ref{fig:SimResultsIncSeg} due to that 
the fact that a causal effect is assumed to exist; this 
reduces the bivariate causal discovery problem to recovering the 
causal ordering over $X_1$ and $X_2$. 
The CD-NOD algorithm cannot easily be extended to 
assume the existence of a causal effect and is therefore
not included in these experiments.

Finally, the results for multivariate causal discovery are presented in 
Figure \ref{fig:SimResultsGeneralCase}, where we plot 
the $F_1$ score 
between 
the true and inferred DAGs as the depth of the mixing-DNN increases. 
The proposed method is competitive across all depths. 
In particular, 
the proposed method 
out-performs the PC algorithm, indicating that  
its use to resolve undirected edges is beneficial. 


\vspace{-1mm}
\subsection{HIPPOCAMPAL FMRI DATA} 
\label{sec_fmriApp}
\vspace{-1mm}

\begin{figure}[t!]
	\begin{center}
		\includegraphics[width=.355\textwidth]{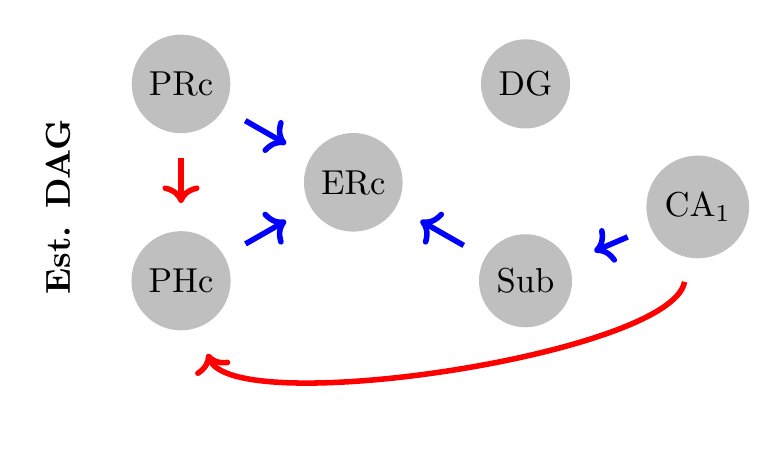}
	\end{center} \vspace{-8mm}
	\caption{Estimated causal DAG on  
		fMRI Hippocampal data by the proposed method. 
		Blue edges are feasible given anatomical connectivity; red edges are not.}
	\label{fmri_res_fig}
\end{figure}

As a real-data application, the proposed method was applied  to resting state 
fMRI data collected from six distinct brain regions as part of the 
MyConnectome project \citep{Poldrack2015}. 
Data was collected from a single subject over 84 successive days.
Further details are provided in Supplementary Material \ref{app_fMRIData}.
We treated each day as a distinct experimental condition and employed the multivariate extension of the proposed method.
For each unresolved edge, 
we employed NonSENS as described in Section \ref{sec_NonSENS_indepTest} with a 5 layer network.
%
%
The results are shown in Figure \ref{fmri_res_fig}. While there is 
no ground truth available, 
we highlight in blue all estimated edges which are 
feasible due to 
anatomical connectivity between the regions and in red estimated edges which are 
not feasible
\citep{Bird2008}.  
We note that the proposed method recovers feasible 
directed connectivity structures for the 
entorhinal cortex (ERc), which is known to play an prominent role within the
hippocampus.

\vspace{-1mm}
\section{CONCLUSION}
\vskip -.2cm
We present a method to perform causal discovery in the context of general non-linear SEMs in the presence of non-stationarities or different conditions. 
This is in contrast to 
alternative methods which often require restrictions on the functional form of the SEMs. 
The proposed method
exploits the correspondence between non-linear ICA  and 
non-linear SEMs, as originally considered in the linear setting by \cite{Shimizu2006}. \kz{Notably, we established the identifiability of causal direction from a completely different angle, by making use of non-stationarity instead of constraining functional classes.  Developing computationally more efficient methods to for the multivariate case is one line of our future work.}

{\newpage 
\bibliographystyle{plainnat}
\bibliography{library.bib}
}

\newpage 
\onecolumn
\section*{Supplementary Material}

\setcounter{equation}{0}
\renewcommand{\theequation}{S.\arabic{equation}}
\setcounter{figure}{0}
\renewcommand{\thefigure}{S.\arabic{figure}}
\renewcommand{\thesection}{\Alph{section}}
\setcounter{section}{0}
\section{Review: Non-linear ICA via time contrastive learning}
\label{sec_TCLreview}

Hwew we provide a more detailed review of the theory of TCL. For further
details we refer readers to \cite{Hyvarinen2016a} and we note that the results presented in 
this section are adapted from \cite{Hyvarinen2016a}.
We begin by showing that an optimally discriminative 
feature extractor combined with a 
linear multinomial classification layer learns to model the 
non-stationary probability density of observations within each 
experimental condition.

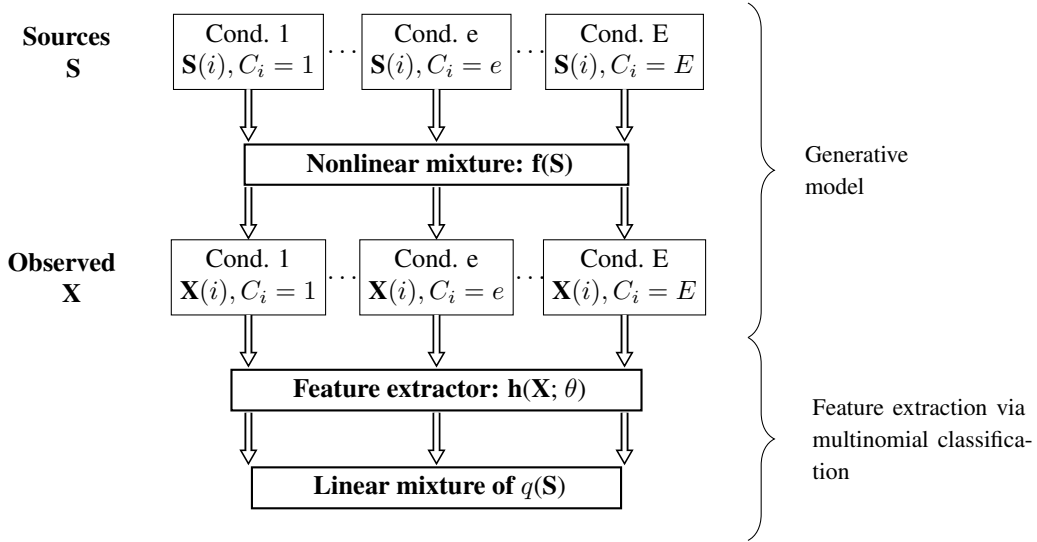
\begin{figure}[h!]
	\centering
	\begin{tikzpicture}
	\node[draw,rectangle, minimum height=1cm, align=center] (0,0) {\mbox{Cond. 1} \\ $\textbf{S}(i), C_i =1$ }; 
	\node at (1.25,0) {$\ldots$}; 
	\node[draw,rectangle, minimum height=1cm, align=center] at (2.5,0) {\mbox{Cond. e} \\ $\textbf{S}(i), C_i =e$}; 
	\node at (3.75,0) {$\ldots$}; 
	\node[draw,rectangle, minimum height=1cm, align=center] at (5.,0) {\mbox{Cond. E} \\ $\textbf{S}(i), C_i =E$}; 
	
	\node [rotate=0, text width=1cm, align=center] at (-2.5, 0) {\textbf{Sources} \\~ ~\textbf{S}};
	
	\draw[vecArrow] (2.5,-.5) to (2.5, -1.2);
	\draw[vecArrow] (0.,-.5) to (0, -1.2);
	\draw[vecArrow] (5,-.5) to (5, -1.2);
	
	\draw[innerWhite] (2.5,-.5) to (2.5, -1.2);
	\draw[vecArrow] (2.5,-1.8) to (2.5, -2.5);
	\draw[innerWhite] (2.5,-1.8) to (2.5, -2.5);
	
	\draw[vecArrow] (0,-1.8) to (0, -2.5);
	\draw[vecArrow] (5,-1.8) to (5, -2.5);
	\node[ draw,  thick, text opacity=1] at (2.5, -1.5) {~~~~~~~~\textbf{Nonlinear mixture:} \textbf{f(S)}~~~~~~~~};  
	
	\node[draw,rectangle, minimum height=1cm, align=center] at (0,-3) {\mbox{Cond. 1}\\ $\textbf{X}(i), C_i =1$}; 
	\node[draw,rectangle, minimum height=1cm, align=center] at (2.5,-3) {\mbox{Cond. e}\\ $\textbf{X}(i), C_i =e$};   
	\node at (1.25,-3) {$\ldots$}; 
	\node at (3.75,-3) {$\ldots$}; 
	\node[draw,rectangle, minimum height=1cm, align=center] at (5.,-3) {\mbox{Cond. E}\\ $\textbf{X}(i), C_i =E$}; 
	
	\node [rotate=0, text width=1cm, align=center] at (-2.7, -3) {\textbf{Observed} \\~ ~ ~ \textbf{X}};
	
	\draw[vecArrow] (2.5,-3.5) to (2.5, -4.2);
	\draw[vecArrow] (0,-3.5) to (0, -4.2);
	\draw[vecArrow] (5,-3.5) to (5, -4.2);
	\draw[innerWhite] (2.5,-3.5) to (2.5, -4.2);
	\draw[vecArrow] (2.5,-4.8) to (2.5, -5.5);
	\draw[vecArrow] (0,-4.8) to (0, -5.5);
	\draw[vecArrow] (5,-4.8) to (5, -5.5);
	\draw[innerWhite] (2.5,-4.8) to (2.5, -5.5);
	\node[ draw,  thick, text opacity=1] at (2.5, -4.5) {~~~~~~~~\textbf{Feature extractor:} $\textbf{h}$(\textbf{X}; $\theta$)~~~~~~~~};
	\node[ draw,  thick, text opacity=1] at (2.5, -5.8) {~~~~~~~~\textbf{Linear mixture of}  $q$(\textbf{S})~~~~~~~~};
	\draw [decorate,decoration={brace,amplitude=10pt,mirror,raise=4pt},yshift=0pt]
	(6.5,-3.8) -- (6.5,0.65) node [black,midway,xshift=1.4cm, text width = 1cm] {\footnotesize
		Generative model};
	\draw [decorate,decoration={brace,amplitude=10pt,mirror,raise=4pt},yshift=0pt]
	(6.5,-6.5) -- (6.5,-3.8) node [black,midway,xshift=2.5cm, text width = 3.cm] {\footnotesize
		Feature extraction via multinomial classification};
	\end{tikzpicture}
	\caption{A cartoon visualization which highlights the relationship between 
		features learnt using TCL and a non-linear ICA model. 
		The non-linear ICA model assumes observations, $\textbf{X}$, are generated based on
		non-stationary latent sources whose distribution varies according the distinct experimental conditions, $e \in \mathcal{E}$.  }
	\label{fig:tcl_cartoon}
\end{figure}

Recall that we observe $d$-dimensional data, $\textbf{X}(i) = \textbf{f}(\textbf{S}(i))$, which is generated 
via a smooth and invertible non-linear mixture, $\textbf{f}$, of $d$ independent latent variables,
$\textbf{S}(i)$. 
As in linear ICA, the latent variables are assumed to be mutually independent. However, we also
assume they are non-stationary. In particular, we assume the 
distribution of latent variables to be piece-wises stationary such that 
we may associate a label, $C_i \in \mathcal{E}$, with each $\textbf{S}(i)$
indicating the piece-wise stationary segment from which it was generated. 
In this manner, it is assumed that the 
distribution of 
latent variables varies across segments, as shown in Figure \ref{fig:tcl_cartoon}. 
%
As such, we write $C_i \in \mathcal{E}$ to denote the segment of the $i$th observation
where $\mathcal{E}=\{1, \ldots, E\}$ is the set of all distinct segments. 
For example, each segment may correspond to a distinct experimental condition. 
As the function $\textbf{f}$ is smooth and invertible, it follows that the distribution of 
each $\textbf{X}(i)$ will also vary across segments.

We may therefore consider the task of classifying observed data into the various segments 
as a multinomial classification task consisting of features, $\textbf{X}(i)$, and
categorical labels, $C_i$. 
For any observation, $\textbf{X}(i)$, associated with true label $C_i \in \mathcal{E}$, 
we have:
\begin{equation}
\label{mlr_posterior_tcl}
p( C_i =  \tau | \textbf{X}(i), \theta, \textbf{W}, \textbf{b}) = \frac{\exp \left ( \textbf{w}_\tau^T ~\textbf{h}( \textbf{X}(i); \theta)~ + b_\tau \right ) }{1 + \sum_{e=2}^E \exp \left ( \textbf{w}_e^T ~\textbf{h}( \textbf{X}(i); \theta)~ + b_e \right ) }, 
\end{equation}
where $\theta$ are parameters for the neural network feature extractor and 
the weight matrix, $\textbf{W} = [\textbf{w}_1, \ldots, \textbf{w}_E]$, and bias vector, $\textbf{b} = [b_1, \ldots, b_E]$, 
parameterize the 
final multinomial layer.
We note that the sum in the denominator goes from $e=2, \ldots, E$. This is because 
we fix $\textbf{w}_1 = 0$ and $b_1=0$ in order to avoid indeterminancy in the softmax function. 

Conversely, we can derive the true posterior distribution over the label $C_i$
as:
\begin{equation}
\label{true_posterior_tcl}
p( C_i = \tau | \textbf{X}(i) )  = \frac{ p(\textbf{X}(i) | C_i = \tau ) ~p(C_i = \tau) }{\sum_{e=1}^E ~p(\textbf{X}(i) | C_i = e ) ~ p(C_i = e)  } .
\end{equation}

If we assume that the feature extractor has a universal function approximation capacity
and that we have infinite data then the multinomial logistic classifier based on 
features $\textbf{h}( \textbf{X}; \theta)$ will converge to the optimal classifier, implying 
that equation (\ref{mlr_posterior_tcl}) will equal equation (\ref{true_posterior_tcl}) for all 
$\tau \in \mathcal{E}$.
We may then consider the following ratio:
\begin{equation}
\frac{ p( C_i =  \tau | \textbf{X}(i), \theta, \textbf{W}, b)}{ p( C_i =  1 | \textbf{X}(i), \theta, \textbf{W}, b)} = \frac{p( C_i = \tau | \textbf{X}(i) )}{p( C_i = 1 | \textbf{X}(i) )},
\end{equation}
which after expanding and taking logarithms yields:
\begin{equation}
\label{tcl_convergence_eq}
\textbf{w}_\tau^T ~ \textbf{h}( \textbf{X}(i); \theta) + b_\tau =  \log ~p(\textbf{X}(i) | C_i = \tau) -
\log ~p(\textbf{X}(i) | C_i = 1) + \log ~ \frac{p(C_i = \tau)}{ p(C_i = 1) } , 
\end{equation}
indicating that the optimal feature extractor computes the log probability density function 
of the data within each experimental condition
(relative to some pivot segment, in this case the first condition). We note that this condition
holds for all $\tau \in \mathcal{E}$.

If we further assume the data were generated according to a non-stationary ICA model 
as described in equation (\ref{gen_model}), then equation 
(\ref{tcl_convergence_eq}) yields:
\begin{equation}
\label{tcl_proof_eq1}
W_\tau^T ~ \textbf{h}( \textbf{X}; \theta) - k_1(\textbf{X}(i))  = \sum_{j=1}^d ~\lambda_{j}(\tau) ~q(S_j) - k_2(\tau) ,
\end{equation}
where the sum is taken over each independent source, $S_1, \ldots, S_d$.
Equation (\ref{tcl_proof_eq1}) follows from the change of variable from $\textbf{S}$ to $\textbf{X}$, noting that the Jacobians required by such a transformation cancel out because of the 
subtraction in the  right hand side of equation (\ref{tcl_convergence_eq}). As a result, by modeling the 
log probability densities with respect to some pivot segment (in this case segment 1), we do not 
need explicity compute the Jacobians. 
We note that  $k_1$ is a function which does not depend on $\tau$ and $k_2$ is a function which does not 
depend on either $\textbf{X}$ or $\textbf{S}$.
As a result, it follows that both 
$\textbf{h}( \textbf{X}; \theta)$ and $q(\textbf{S})$ must span the same linear space, implying that 
we may compute 
latent sources up to some non-linearity, $q(\textbf{S})$, 
by first learning a feature extractor based on TCL and 
subsequently applying linear ICA on estimated features, $\textbf{h}( \textbf{X}; \theta)$. 
Figure \ref{fig:tcl_cartoon} summarizes the relationship between the generative model and
TCL. 

\section{A linear ICA method for piece-wise stationary sources}
\label{supp_linearICA_SM}
Formally, assumptions 1-3 of Theorem \ref{theorem_tcl} guarantee that  TCL, as presented in 
Section \ref{sec_nonLinICA}, will 
recover a linear mixture of latent independent sources up to point-wise transformation. 
This implies that the hidden representations obtained satisfy:
\begin{equation}
\label{TCL_lastICA_step_eq}
\textbf{h}(\textbf{X}; \theta) = \textbf{A}  q(\textbf{N}),
\end{equation}
for some linear mixing matrix, $\textbf{A}$. Equation (\ref{TCL_lastICA_step_eq})
suggests that applying ordinary linear ICA to hidden representations, $\textbf{h}(\textbf{X};\theta)$,
will allow us to recover $q(\textbf{N})$. However, the use of linear ICA 
is premised on the assumption that 
latent 
variables 
are independent. This is not necessarily 
guaranteed under the generative model presented in equation (\ref{gen_model}) in the context of a
fixed 
number of segments.
While it is certainly true that 
$N_1$ and $N_2$ are conditionally independent given segment labels, it may be the case 
that they are not marginally independent over all segments; for example it is possible that their variances 
are somehow related (e.g., they may be monotonically increasing). 
We note that this is not an issue when the number of segments grows asymptotically and we
assume exponential family parameters, $\lambda_j(e)$, are randomly generated, as stated
in Corollary 1 of \cite{Hyvarinen2016a}.

Here we seek to address this issue by proposing an alternative linear ICA algorithm which explicitly
models the piece-wise stationary nature of the data. In particular, the proposed linear ICA 
algorithm explicitly incorporates equation (\ref{gen_model}) as the generative model for latent sources.
As such, it can be used to accurately unmix sources in the final stage of TCL, especially when the
number of segments in small. 

To set notation, we assume we observe data which corresponds to a linear mixture of sources:
\begin{equation*}
\textbf{Z} = \textbf{A} \textbf{S}
\end{equation*}
where $\textbf{Z}, \textbf{S} \in \mathbb{R}^d$ and 
 $\textbf{A}\in\mathbb{R}^{d \times d}$ 
is a square mixing matrix.
Popular ICA algorithms, such as FastICA, estimate the unmixing matrix $\textbf{W} = \textbf{A}^{-1}$
by maximizing the log-likelihood under the assumptions that 
sources are independent and non-Gaussian. 
The objective function for FastICA is therefore of the form:
	\begin{equation}
\label{FastICA_obj}
\log p(\textbf{Z}) = \sum_{j=1}^d q( \textbf{w}_j^T \textbf{Z}) -   Z(W) ,
\end{equation}
where $Z(W) = \log | \det W |$ is the normalization constant and we write $\textbf{w}_k$ to denote
the $k$th row of $\textbf{W}$. 
Typically a parametric form is assumed for $q(\cdot)$, popular examples being exponential or log-cosh. 
In the context of the generative model for sources specified in equation (\ref{gen_model}),
the FastICA algorithm therefore proceeds under the assumption that $\lambda_j(e)$ is constant across all
segments $e \in \mathcal{E}$. 

In order to address this issue we consider an alternative model for the density of latent sources. In particular, we seek to directly employ the piece-wise stationary log-density 
described in equation (\ref{gen_model}). As such, we log-density of an observation within segment 
$e \in \mathcal{E}$ is defined as:
	\begin{equation}
\label{piecewiseICA_eq}
\log p_e( \textbf{Z}) = \sum_{j=1}^d \lambda_j (e)  q_j( \textbf{w}_j^T \textbf{Z}) -  Z(W, \lambda(e)).
\end{equation}
In contrast to equation (\ref{FastICA_obj}), the log-density of each observation 
depends on both the segment, $e$, the exponential family parameters, 
$\boldsymbol \lambda = \{ \lambda_j(e): e \in \mathcal{E}, j =1, \ldots, n \}$,
as well as the unmixing matrix, $\mathbf{W}$. 

In other to recover latent sources we propose to estimate parameters, corresponding to 
unmixing matrix as well as exponential family parameters, 
via score matching \citep{Hyvarinen2006}. This avoids the need to estimate the 
normalization parameter, which may not be available analytically when sources follow
unnormalized distributions.
The score matching objective
for the ICA model defiend in equation (\ref{piecewiseICA_eq}) is defined as:
\begin{align}
\label{SM_obj}
\begin{split}
\tilde J &= \sum_{e \in \mathcal{E}} \sum_{j=1}^d \lambda_j (e) \frac{1}{n_e} ~\sum_{i, C_i=e} q_j'' (\textbf{w}^T_j \textbf{Z}(i) ) \\&+ 
\frac{1}{2} \sum_{e \in \mathcal{E}} \sum_{j,k=1}^d \lambda_k(e) \lambda_j(e) \textbf{w}_k^T \textbf{w}_j
\frac{1}{n_e} \sum_{i, C_i=e} q_k'(\textbf{w}^T_k \textbf{Z}(i)) q_j'(\textbf{w}^T_j \textbf{Z}(i) ),
\end{split}
\end{align}
where we write $q_j'$ and $q_j''$ to denote the first and second derivatives of 
$q_j$ with respect to observations, $\textbf{Z}$. 
We propose to minimize equation (\ref{SM_obj}) via block gradient descent, conditionally updating 
the mixing matrix $\textbf{W}$ and exponential family parameters $\boldsymbol {\lambda}$. This has the important 
benefit that conditional on $\textbf{W}$, there is a closed form update for 
$\boldsymbol \lambda$ \citep{Hyvarinen2007}.



\subsubsection*{Experimental results}
In order to assess the performance of the proposed linear ICA algorithm, we 
generated bivariate data following the piece-wise stationary distribution described in 
equation (\ref{gen_model}). 
We compare the performance of the proposal algorithm against 
the following popular linear ICA algorithms: FastICA \citep{Hyvarinen1999}, 
Infomax ICA  \citep{bell1995information} and 
Joint Diagonalization method proposed by 
\cite{Pham2001a}
which also accommodates non-stationary 
sources. 

In order to assess the performance of the proposed method we consider two scenarios: 
\begin{itemize}
	\item The exponential family parameters are deliberately generated 
	such that there is a statistical dependence structure across segments. In particular, we 
	generate bivariate data where we explicitly enforce 
	$\lambda_j(e)$
	to be monotonically increasing in 
	$e$. 
	As a concrete example, when sources follow a 
	Laplace distribution this implies that $q(S) = |S|$ and in turn $\lambda_j(e)$
	corresponds to the variance of the $j$th source in segment $e$. 
	In such a setting, we generate piece-wise stationary Laplace sources with where the variances 
	are correlated across segments. 
	\item As a baseline, we also generate data where the exponential family parameters are 
	generated at random. This removes any systematic, higher-order 
	dependence between latent sources.

\end{itemize}

\begin{figure*}[h!]
	\centering
	\includegraphics[width=.65\textwidth]{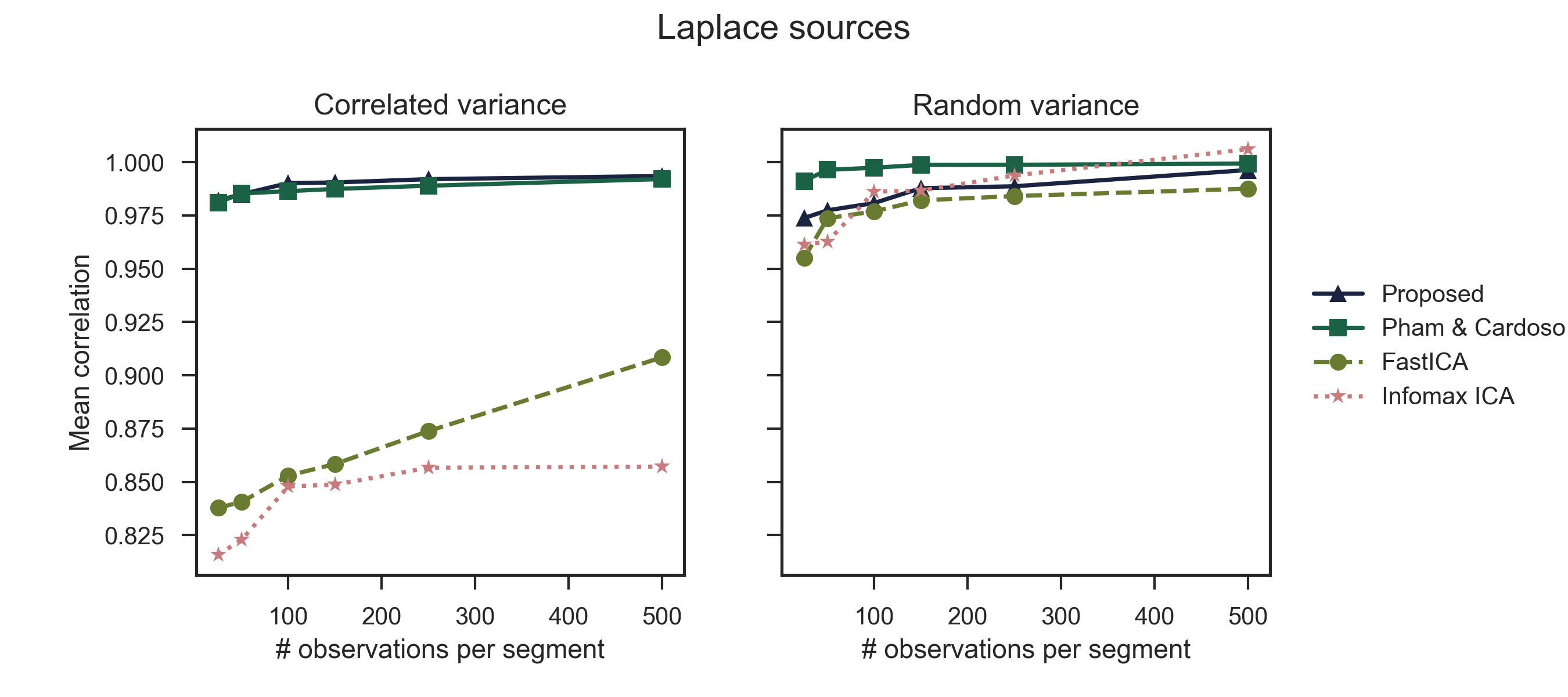}
	\caption{Performance of various linear ICA algorithms for piece-wise stationary 
	Laplace sources. The left panel shows performance when 
	the variance of latent sources are positive correlated, introducing second order dependence 
	in the data. As expected, FastICA and Infomax ICA perform poorly in this context. 
	The right panel shows similar data were variances are no longer correlated, resulting in 
	good performance for all algorithms.  }
	\label{fig:LapSources_linICA}
\end{figure*}

We begin by generating bivariate data were sources follow a piece-wise stationary 
Laplace distribution. This implies that 
sources follow the log-density specified in equation (\ref{gen_model})
where $q_j(S_j) = |S_j|$ and each term $\lambda_j(e)$ denotes the variance of the $j$th source
in segment $e$. 
Results when data is generated over five segments are provided in Figure \ref{fig:LapSources_linICA}.
The left panel shows the case were the variances of each latent source are correlated across segments.
We note that when this is the case, methods such FastICA and Infomax ICA perform poorly.
This is in contrast to the proposed method and the joint diagonalization 
approach of \cite{Pham2001a}, who explicitly model the non-stationary 
nature of the data. 
The right panel of Figure \ref{fig:LapSources_linICA} shows equivalent results when variances are 
randomly generated, thereby removing 
second order dependence between latent variables. As expected, in this setting all methods perform
well. 

\begin{figure*}[h!]
	\centering
	\includegraphics[width=.65\textwidth]{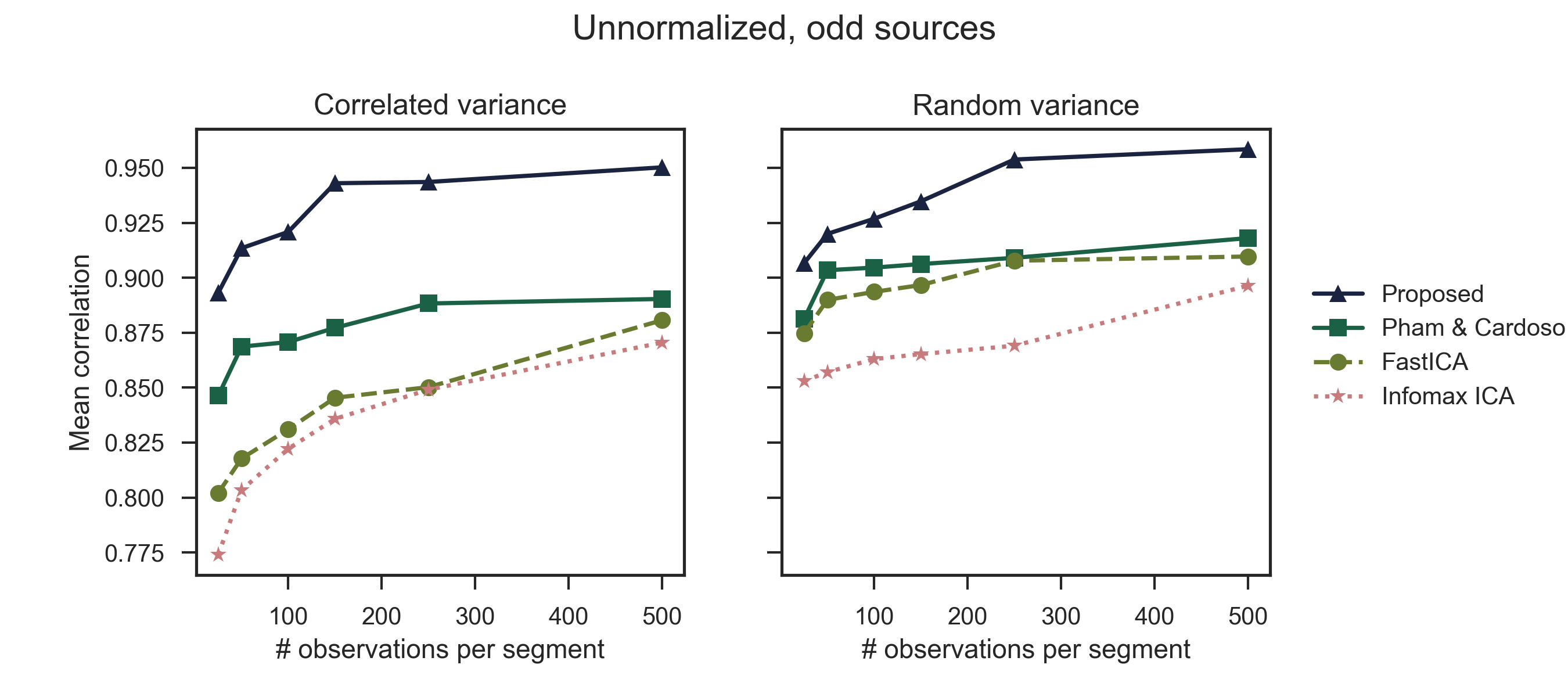}
	\caption{Performance of various linear ICA algorithms when latent sources
		follow the log-density detailed in equation (\ref{logDensityEq}). The left panel shows performance when 
		exponential family parameters, $\lambda_i(e)$, 
		are positive correlated, introducing higher order dependence 
		in the data. The right panel shows similar data were parameters, $\lambda_i(e)$
		are randomly generated. }
	\label{fig:OddSources_linICA}
\end{figure*}

In order to further probe the differences between the proposed method and 
the approach of \cite{Pham2001a}, we consider 
latent sources with an unnormalized distribution. In particular, we generate 
sources 
such that the log-density within a particular segment is as follows:
	\begin{equation}
\label{logDensityEq}
\log p_e( S_j) = 
\begin{cases}
-3 \lambda_j(e) |S_j| - \frac{1}{2} S_j^2 ,& \text{if } S_j\geq 0\\
-\lambda_j(e) |S_j| - \frac{1}{2} S_j^2,              & \text{otherwise}.
\end{cases}
\end{equation}
Such a density is both unnormalized but also odd. As such, we expect the joint diagonalization algorithm
of \cite{Pham2001a} to perform poorly in this setting as it exclusively studies covariance 
structure and therefore cannot model skewed distributions. 
Figure \ref{fig:OddSources_linICA} visualizes the results for these experiments. As expected, 
the Joint Diagonalization algorithm of \cite{Pham2001a} 
suffers a drop in performance. However, it continues to outperform 
FastICA and Infomax ICA, especially when there are dependencies over the 
exponential family parameters.
We note that the proposed model, where parameters are estimated using score matching,
is shown to be more robust.

\section{Proof of Theorem \ref{TCL_causalRecov_prop}}
\label{supp_Proof_Theorem2}
The proof of Theorem \ref{TCL_causalRecov_prop} follows from a combination of the 
presented assumptions together with Property \ref{Prop_DAG_indep}. 
Formally, assumptions 1--3 guarantee that the TCL, as presented in 
Section \ref{sec_nonLinICA}, will 
recover a linear mixture of latent independent sources up to point-wise transformation. 
This, combined with the novel linear ICA algorithm described in Section \ref{SM_linICAmodel},
imply that latent disturbances can be recovered. 
We note that in practice 
we recover the latent disturbances up to 
point-wise transformation, $q( \textbf{N})$, as opposed to $\textbf{N}$. 
This is not a problem when a general test of statistical independence, such as HSIC
which can capture arbitrary (i.e., non-linear)  dependencies, is employed. 
Finally, 
Assumption 1 further states there is no latent confounder is present, implying that 
running all possible pairwise tests 
using a sufficiently flexible 
independence test, as required by assumption 4, 
will allow us to 
determine causal structure.

\section{Relationship between likelihood ratio and measures of independence}
\label{LR_to_indTest_supp}
In this section we derive the  result presented in equation (\ref{LR_ind_eq}), for some permutation $\pi$ of latent disturbances. 
We begin by considering the mutual information between $X_1$ and $N_{\pi(2)}$:
\begin{align*}
    I( X_1, N_{\pi(2)}) &= H(X_1) + H(N_{\pi(2}) - H(X_1, N_{\pi(2)})\\
    &= H(X_1) + H(N_{\pi(2}) - H(X_1, X_2) -  \mathbb{E} \left [\log \left | \frac{\partial \textbf{g}_{\pi(2)} }{\partial  X_2}\right | \right ]
\end{align*}
where we employ the same change of variable, whose Jacobian can be easily evaluated, as in 
Section \ref{sec_assumeCause}. 
In particular, we have used the property that:
\begin{equation*}
    H(X_1, N_{\pi(2)})  = H(X_1, X_2) + \log | \mbox{det} ~\mathbf{ J \tilde g}|
\end{equation*}
and noted that the particular choice of $\mathbf{\tilde g}$ allows us to directly compute the Jacobian as $\frac{\partial \textbf{g}_{\pi(2)} }{\partial  X_2}$. 
We may therefore compute the difference in mutual information between each observed variable and the relevant
latent disturbance, yielding:
\begin{align*}
     I( X_1, N_{\pi(2)})  -  I( X_2, N_{\pi(1)}) &= 
     H(X_1) + H(N_{\pi(2}) -   \mathbb{E} \left [\log \left | \frac{\partial \textbf{g}_{\pi(2)} }{\partial  X_2}\right | \right ] \\
     &- H(X_2) - H(N_{\pi(1}) +  \mathbb{E} \left [\log \left | \frac{\partial \textbf{g}_{\pi(1)} }{\partial  X_1}\right | \right ]
\end{align*}
which is precisely the negative of likelihood ratio 
presented in Section \ref{sec_assumeCause}.

\newpage 
\section{Baseline methods}
\label{app_baselines}
In this section we briefly overview and provide pseudo-code for alternative methods which are presented as baselines in the 
manuscript. 

\begin{itemize}
	\item \textbf{DirectLiNGAM}:	
	The DirectLiNGAM method of \cite{Shimizu2011} is based on the property that 
	within a linear non-Gaussian acyclic model (LiNGAM), if we regress out the 
	parents of any variable, then the residuals also follow a LiNGAM. 
	Based on this property, 
	the authors propose an iterative algorithm through which to iteratively 
	uncover exogenous variables.
	o
	Further, if variables follow a LiNGAM, then due to the additive nature of noise in such models,  the residuals will be independent when we 
	regress the parent on its children. As a result, we may infer the causal structure by studying 
	the statistical independences between variables and 
	residuals. 
	Pseudo-code for the bivariate DirectLiNGAM method is provided in Algorithm \ref{alg_directLiNGAM}.
	
	\begin{algorithm}[h]
		\DontPrintSemicolon
		\SetAlgoLined
		\SetKwInOut{Input}{Input}
		\SetKwInOut{Output}{Output}
		\Input{Bivariate data, $\textbf{X}, $ 
			and significance level $\alpha$.}
		\For{$i \in \{1,2\} $ }{
			Linearly regress $X_i$ on $X_{ \{1,2\} \backslash i}$ and compute the residual $\hat N_i$.\; 
			Evaluate the test:		
			\begin{align*}
			H_{X_i, \hat N_i, 0}:  \textbf{P}_{X_i, \hat N_j} = \textbf{P}_{X_i} \textbf{P}_{\hat N_j} ~~\mbox{against}~~ H_{X_i, \hat N_j, 1}: \textbf{P}_{X_i, N_j} \neq \textbf{P}_{X_i}\textbf{P}_{\hat N_j} ~~\mbox{at the $\frac{\alpha}{2}$ level.}
			\end{align*}
		}
		\uIf{we fail to reject the null hypothesis only once  }{
			Variable $i'$ such that we fail to reject $H_{X_{i'}, \hat N_j, 0}$ is the cause variable
		}\Else{
			The causal dependence structure is inconclusive 
		}
		\caption{Bivariate causal discovery using DirectLiNGAM \hfill  \citep{Shimizu2011}}
		\label{alg_directLiNGAM}
	\end{algorithm}

	\item \textbf{RESIT}:
	The RESIT method, first proposed by \cite{Hoyer2009} and subsequently extended 
	by \cite{Peters2013}, 
	can be seen as a non-linear extension of the DirectLiNGAM algorithm. 
	The RESIT algorithm is premised on the assumption of an
	additive noise model (ANM), which implies that 
	each structural equation are of the form $X_j = f_j(\textbf{PA}_j) + N_j$. 
	Given the ANM assumption, RESIT is able to recover the causal structure by 
	testing for dependence 
	between variables and residuals. 
	Gaussian process regression is employed in order to accommodate for 
	non-linear additive causal relations.
	\begin{algorithm}[h]
		\DontPrintSemicolon
		\SetAlgoLined
		\SetKwInOut{Input}{Input}
		\SetKwInOut{Output}{Output}
		\Input{Bivariate  data, $\textbf{X}, $ 
			and significance level $\alpha$.}
		\For{$i \in \{1,2\} $ }{
			Regress $X_i$ on $X_{ \{1,2\} \backslash i}$ and compute the residual $\hat N_i$. \tcp*{Gaussian process regression is employed}
			Evaluate the test:		
			\begin{align*}
			H_{X_i, \hat N_i, 0}:  \textbf{P}_{X_i, \hat N_j} = \textbf{P}_{X_i} \textbf{P}_{\hat N_j} ~~\mbox{against}~~ H_{X_i, \hat N_j, 1}: \textbf{P}_{X_i, N_j} \neq \textbf{P}_{X_i}\textbf{P}_{\hat N_j} ~~\mbox{at the $\frac{\alpha}{2}$ level.}
			\end{align*}
		}
		\uIf{We fail to reject the null hypothesis only once  }{
			Variable $i'$ such that we fail to reject $H_{X_{i'}, \hat N_j, 0}$ is the cause variable
		}\Else{
			The causal dependence structure is inconclusive 
		}
		\caption{Bivariate causal discovery using RESIT \hfill \citep{Hoyer2009, Peters2013}}
		\label{alg_RESIT}
	\end{algorithm}

	\item \textbf{Non-linear ICP}:	The ICP method proposed by \cite{Peters2016} proposes a fundamentally different approach to 
	causal discovery. 
	The underlying principle of the ICP algorithm is that the direct causal predictors of a
	given variable must remain invariant across distribution shifts induced by various 
	experimental conditions. 
	In the context of bivariate data, the non-linear ICP algorithm, as described in 
	Section 6.1 of \cite{Peters2016} therefore corresponds to fitting a 
	non-linear regression model on the data across all experimental conditions and
	testing whether the distribution of residuals is the same within each condition.  
	We note that such an approach assumes an additive noise model, as this greatly simplifies testing for 
	invariance.
	\begin{algorithm}[h]
		\DontPrintSemicolon
		\SetAlgoLined
		\SetKwInOut{Input}{Input}
		\SetKwInOut{Output}{Output}
		\Input{Bivariate data, $\textbf{X}(i) $, labels $C_i \in \{1, \ldots, E\}$
			and significance level $\alpha$.}
		\For{$i \in \{1,2\} $}{
			Regress $X_i$ on $X_{ \{1,2\} \backslash i}$ and compute the residual, $\hat N_i$\; 
			\tcp{Note that Gaussian process regression is employed and we aggregate data across all experimental conditions}
			Evaluate the test:
			\begin{align*}
			H_{\hat N_i, 0}:  \textbf{P}_{\hat N_i, e} = \textbf{P}_N ~\mbox{for all $e \in \mathcal{E} $} ~~\mbox{against}~~
			H_{\hat N_i, 1}:  \textbf{P}_{\hat N_i, e} \neq \textbf{P}_N ~\mbox{for some $e \in \mathcal{E}$}
			\end{align*}
			where $\textbf{P}_N$ is some arbitrary distribution. \tcp{The Kolmogorov-Smirnov test is employed}
		}
		\uIf{$ \text{We fail to reject the null hyp. only once} $}{
			Variable $i$ such that we fail to reject $H_{\hat N_i, 0}$  is the cause variable
		}\Else{
			The causal dependence structure is inconclusive 
		}
		\caption{Bivariate causal discovery via non-linear ICP \hfill \citep{Peters2016}}
		\label{my_alg}
	\end{algorithm}
	
	\item \textbf{RECI}:	\cite{Blobaum2018} propose a method for inferring the causal relation between two variables
	by comparing the regression errors in each possible causal direction. 
	Under some mild assumptions, they are able to prove that the 
	magnitude of residual errors should be smaller in the causal direction. This suggests a
	straightforward causal discovery algorithm, which we outline below. We note that while any non-linear regression method may be employed, our implementation used Gaussian process regression.
	
	\begin{algorithm}[h]
		\DontPrintSemicolon
		\SetAlgoLined
		\SetKwInOut{Input}{Input}
		\SetKwInOut{Output}{Output}
		\Input{Bivariate data, $\textbf{X} $.}
		Standardize data such that $\textbf{X}_t$ is zero mean and unit variance.\;
		\For{$i \in \{1,2\} $ }{
			Regress $X_i$ on $X_{ \{1,2\} \backslash i}$ and evaluate the mean-squared error, $MSE_i$ 
			\tcp{Gaussian process regression is employed} 
		}
		\uIf{ $MSE_1 < MSE_2$}{
			Variable 1 is the cause variable
		}\Else{
			Variable 2 is the cause variable
		}
		\caption{Bivariate causal discovery using RECI \hfill \citep{Blobaum2018}}
		\label{alg_RECI}
	\end{algorithm}
	
	\item \textbf{CD-NOD}:	
	\cite{Zhang2017a} propose a causal discovery algorithm which explicitly 
	accounts for non-stationarity or heterogeneity 
	over observed variables. 
	The CD-NOD algorithm
	accounts for non-stationarity, which may manifest itself as changes in the causal modules,
	by  introducing an additional variable 
	representing the time or domain index into the causal DAG.
	Conditional independence testing is then employed to 
	recover the skeleton over the augmented DAG. 
	Their method can find causal direction by making use of not only invariance, but also independent changes of causal models, as an extended notion of invariance. 
	We also note that CD-NOD is a non-parametric method, implying it is able to accommodate 
	non-linear causal dependencies. 
	
	\begin{algorithm}[h]
		\DontPrintSemicolon
		\SetAlgoLined
		\SetKwInOut{Input}{Input}
		\SetKwInOut{Output}{Output}
		\Input{Bivariate  data, $\textbf{X} $.}
		Build an augmented dataset consisting of $\textbf{X}$ and $C$, the observed variable representing time or
		domain index.\;
		\tcp{Detection of changing modules}
		\For{$i \in \{1,2\} $ }{
			Test for marginal and conditional dependence between $X_i$ and $C$\; 
			If they are conditionally
			independent given $X_{\{1,2\} \backslash i}$ then we remove the edge between 
			$X_i$ and $C$ in the augmented DAG\;
		}
		\tcp{Recover causal skeleton}
		\uIf{$X_1 \bigCI X_2 ~ | ~ C $ }{
			Remove the edge between $X_1$ and $X_2$ $\Rightarrow$  no causal relation between $X_1$ and $X_2$
		}
		\Else{
			\uIf{Only one of $X_1$ and $X_2$ is marginally or conditionally dependent on $C$}{
				Dependent variable is reported as the cause
			}
			\Else{Determine cause variable by comparing mutual information}
		} 
		\caption{Bivariate causal discovery using CD-NOD \hfill \citep{Zhang2017a}}
		\label{alg_CDNOD}
	\end{algorithm}

	
	
\end{itemize}

\newpage 
\section{Further experimental results}
\label{extra_experiments_supp}

In this section of the supplementary material we present further
experimental results. In particular, in Section 
\ref{extra_bivariate_sims} we 
present results for bivariate causal discovery in the context of a fixed number of 
experimental conditions, $|\mathcal{E}|=10$, and increasing observations 
per segment. 
In Section \ref{sec::HammingDist} we provide addition 
results in the context of multivariate causal discovery.
In particular, we report the 
Hamming distance between true and estimated DAGs.

\subsection{Additional bivariate causal discovery experiments}
\label{extra_bivariate_sims}

\begin{figure*}[h!]
	\centering
	\includegraphics[width=.9\textwidth]{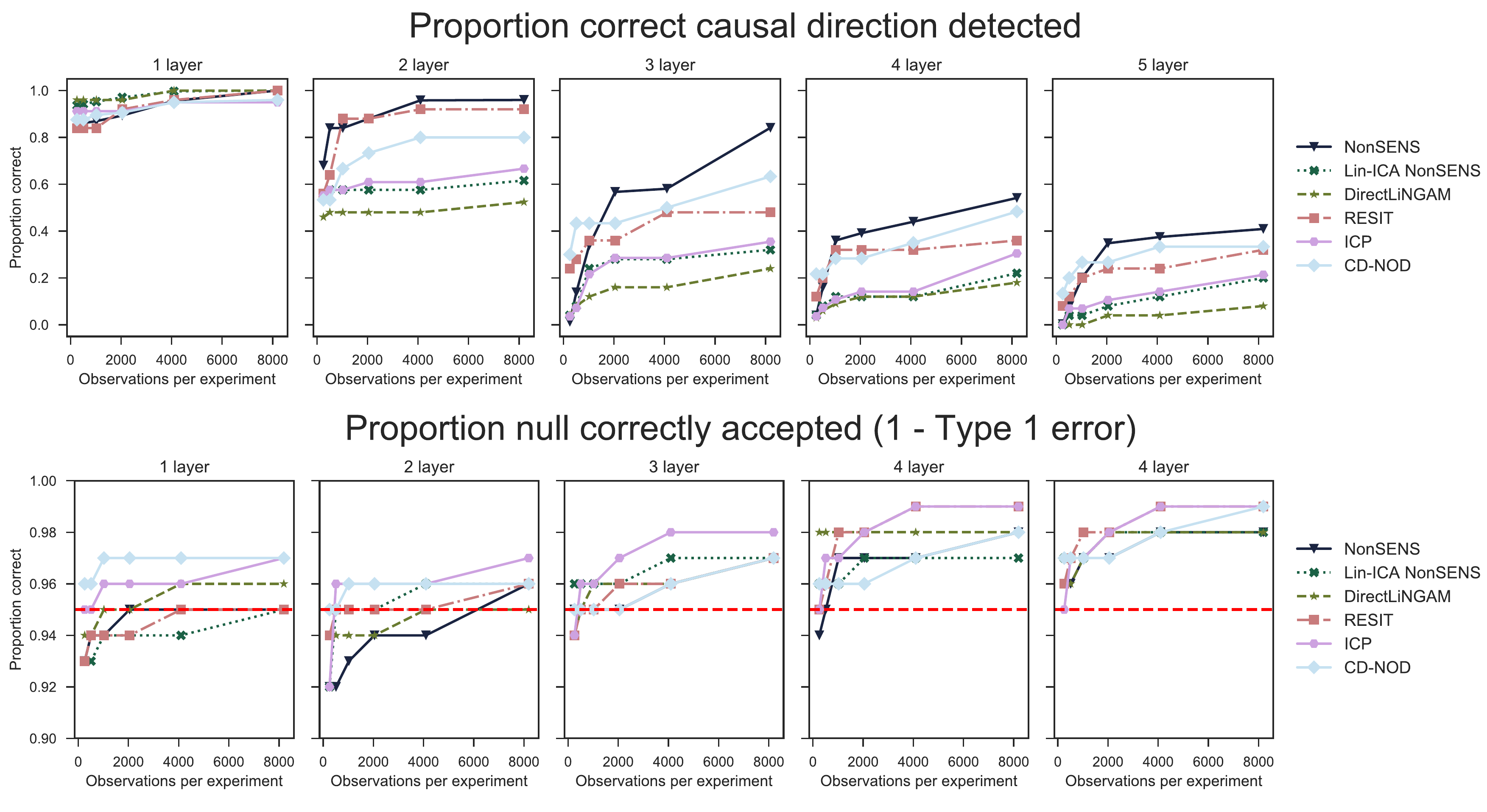}
	\caption{Experimental results 
		indicating performance as we increase the number of observations, $n_e$ conditions,
		within each experimental condition for a fixed number of experimental conditions, $|\mathcal{E}|=10$. 
		Each horizontal plane plots results for varying depths of the mixing-DNN, ranging from $l=1, \ldots, 5$.
		The top panel plots the proportion of times the correct cause variable is identified when a causal effect exists. The bottom panel considers data where no 
		acyclic causal structure exists
		($\textbf{A}^{(l)}$ is not lower-triangular) and reports the 
		proportion of times no causal effect is correctly reported. 
		The dashed, horizontal 
		red line indicates the theoretical $(1-\alpha)\%$ true negative rate. 
		For clarity we omit the standard errors, but we note that they were  small in magnitude (approximately $2-5\%$). 
	}
	\label{fig:SimResultsIncObsPerSeg}
\end{figure*}

We  consider the performance of all algorithms in the context of 
a fixed number of 
experimental conditions, $|\mathcal{E}|=10$, 
and an increasing number of observations per condition, $n_e$. 
The results  are presented in 
Figure \ref{fig:SimResultsIncObsPerSeg}, where 
we repeat each experiment 100 times. 
We note that all algorithms are able to accurately identify 
causal structure in the presence of LiNGAMs (corresponding to a 1 layer mixing-DNN). 
However, as the causal structure becomes increasingly non-linear, the performance of all 
methods declines. 
In particular, we note that the proposed method has comparable performance with 
alternative methods such as RESIT and CD-NOD when the number of samples is small.
However, as the 
number of observations increases
the proposed method is able to out-perform alternative algorithms. 

\subsection{Multivariate causal discovery results}
\label{sec::HammingDist}

\begin{figure}[h!]
	\centering
	\includegraphics[width=.75\textwidth]{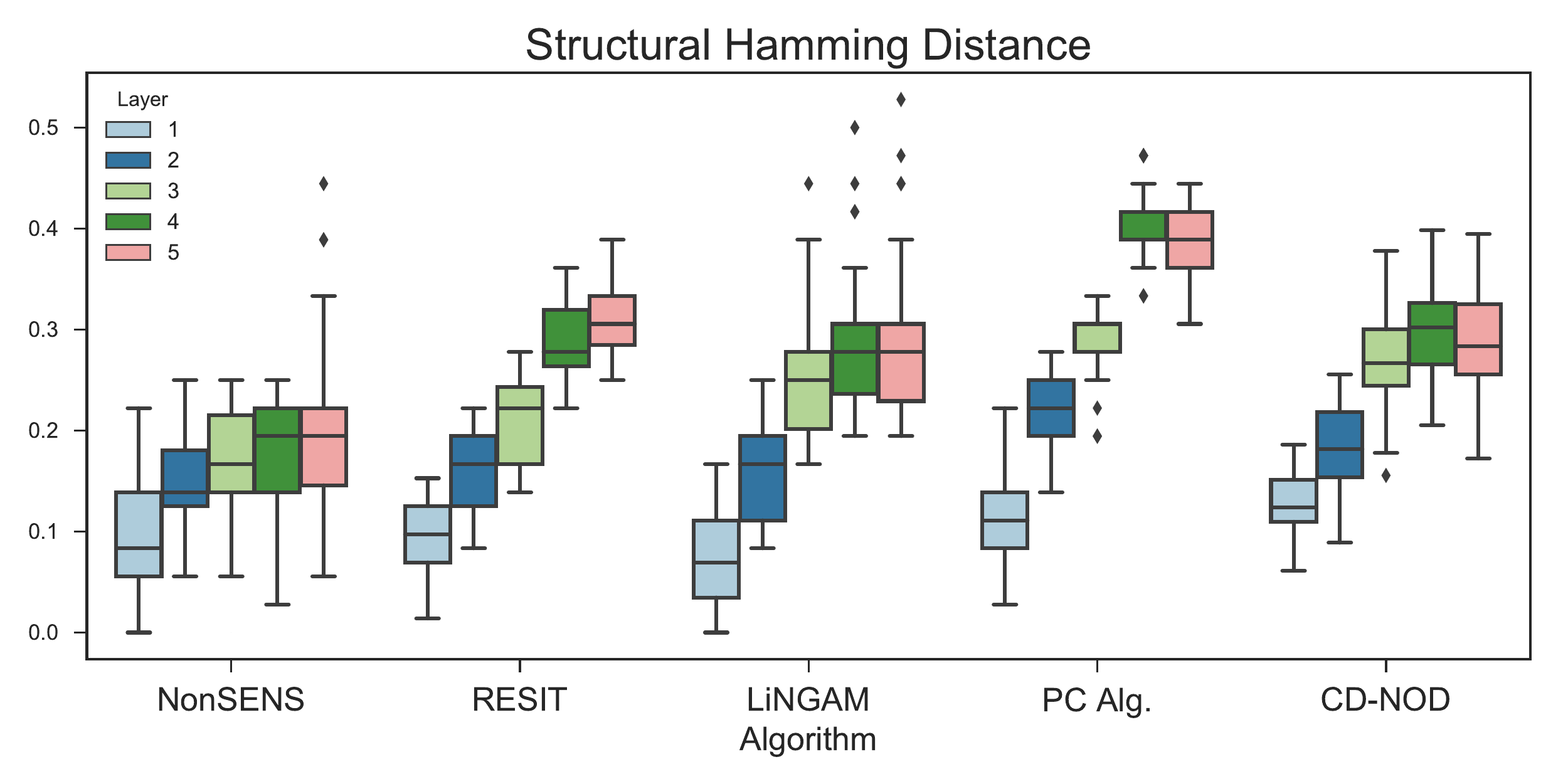}
	\caption{Hamming distance  results for multivariate causal discovery with 
		$6$-dimensional data. 
		For each algorithm, we plot the structural Hamming distance as we vary the 
		depth of the mixing-DNN from $l=1, \ldots, 5$.
		Lower scores indicate better performance.
	}
	\label{fig:SimResultsGeneralCase_sup}
\end{figure}

In this section we provide additional performance metrics in the context of 
multivariate causal discovery. While Figure \ref{fig:SimResultsGeneralCase} 
reported the 
$F_1$ score, we further provide  results for the Hamming distance between true and estimated 
DAGs in Figure \ref{fig:SimResultsGeneralCase_sup}.

.
\newpage 
.
\section{Hippocampal functional MRI data}
\label{app_fMRIData}

In this section we provide further details of the Hippocampal fMRI data employed in Section 
\ref{sec_fmriApp}. 
The data was collected as part of the 
MyConnectome project, presented in \cite{Poldrack2015},
which involved 
daily fMRI scans for a single individual (Caucasian male, aged 45).
The data may be freely downloaded from \verb+https://openneuro.org/datasets/ds000031/+.

We focus only on the resting-state fMRI data taken from this project, noting that 
future work may also wish to study the other modalities of data collected.

Data was collected from the same subject over a series of 84 successive days,
allowing us to consider data collected on distinct days as a distinct experimental condition.
Full details of the data acquisition pipelines are provided in 
\cite{Poldrack2015}. 
For each day, we observe 518 BOLD measurements. 
After preprocessing, data was collected from the following brain regions:
perirhinal cortex (PRC), parahippocampal cortex (PHC), entorhinal cortex (ERC), subiculum (Sub), CA1, and CA3/Dentate Gyrus (DG).
This resulted in $d=6$ dimensional data. 
%
As such, data employed consists of $n_e=518$ observations per experimental condition 
and $|\mathcal{E}| = 84$ distinct conditions.

\end{document}